\documentclass{article}

\usepackage{arxiv_style}

\usepackage{comment}
\usepackage{subcaption}
\usepackage{bookmark}
\usepackage{float}
\usepackage{placeins}
\usepackage{authblk}        
\usepackage{booktabs}       
\usepackage{amsfonts}       
\usepackage{amsmath}        
\usepackage{amssymb}        
\usepackage{graphicx}       
\usepackage{wrapfig}
\usepackage{xcolor}
\usepackage{mdframed}
\usepackage[
    backend=biber,
    style=nejm,
]{biblatex}

\date{}

\renewcommand\cite{\parencite}

\begin{document}
\title{\centering\huge{AI-based Anomaly Detection for Clinical-Grade Histopathological Diagnostics}}

\author[1,2]{Jonas~Dippel*}
\author[3,4]{Niklas~Prenißl*}
\author[1,2]{Julius~Hense†}
\author[5]{Philipp~Liznerski†}
\author[6]{Tobias~Winterhoff}
\author[3]{Simon~Schallenberg}
\author[5]{Marius~Kloft}
\author[7]{Oliver~Buchstab}
\author[3,8]{David~Horst}
\author[3,6]{Maximilian~Alber}
\author[6]{Lukas~Ruff‡}
\author[1,2,9,10]{Klaus-Robert~Müller‡}
\author[2,3,7,8]{Frederick~Klauschen‡}

\affil[1]{Machine Learning Group, Technische Universität Berlin, Berlin, Germany.}
\affil[2]{{BIFOLD – Berlin Institute for the Foundations of Learning and Data, Berlin, Germany.}}
\affil[3]{{Institute of Pathology, Charité – Universitätsmedizin Berlin, Berlin, Germany.}}
\affil[4]{{Berlin Institute of Health at Charité – Universitätsmedizin Berlin, BIH Biomedical Innovation Academy, BIH Charité Junior Digital Clinician Scientist Program, Charitéplatz 1, 10117 Berlin, Germany.}}
\affil[5]{{RPTU, Kaiserslautern, Germany.}}
\affil[6]{Aignostics GmbH, Berlin, Germany.}
\affil[7]{{Institute of Pathology, Ludwig-Maximilians-Universität, Munich, Germany.}}
\affil[8]{{German Cancer Research Center (DKFZ) \& German Cancer Consortium (DKTK), Munich Partner Site.}}
\affil[9]{{Department of Artificial Intelligence}, {Korea University, Seoul, Korea.}}
\affil[10]{{Max-Planck Institute for Informatics}, {Saarbr\"ucken, Germany.}}
\affil[*†‡]{Equal contribution.}

\maketitle

\begin{abstract}
While previous studies have demonstrated the potential of AI to diagnose diseases in imaging data, clinical implementation is still lagging behind. This is partly because AI models require training with large numbers of examples only available for common diseases. In clinical reality, however, only few diseases are common, whereas the majority of diseases are less frequent (long-tail distribution). Current AI models overlook or misclassify these diseases. We propose a deep anomaly detection approach that only requires training data from common diseases to detect also all less frequent diseases.
We collected two large real-world datasets of gastrointestinal biopsies, which are prototypical of the problem. Herein, the ten most common findings account for approximately 90\% of cases, whereas the remaining 10\% contained 56 disease entities, including many cancers. 17 million histological images from 5,423 cases were used for training and evaluation. Without any specific training for the diseases, our best-performing model reliably detected a broad spectrum of infrequent (``anomalous'') pathologies with 95.0\% (stomach) and 91.0\% (colon) AUROC and generalized across scanners and hospitals.
By design, the proposed anomaly detection can be expected to detect \emph{any} pathological alteration in the diagnostic tail of gastrointestinal biopsies, including rare primary or metastatic cancers.
This study establishes the first effective clinical application of AI-based anomaly detection in histopathology that can flag anomalous cases, facilitate case prioritization, reduce missed diagnoses and enhance the general safety of AI models, thereby driving AI adoption and automation in routine diagnostics and beyond.
\end{abstract}


\newpage

\section{Introduction} 

Diagnostic pathology is facing serious challenges due to a shortage of pathologists in many parts of the world and too few young doctors entering the profession \cite{pathologists-gap}.
Meanwhile, the diagnostic workload, and cancer burden in particular, is rising under an aging population \cite{cancercases}. Moreover, diagnostic procedures are getting more complex due to the demands of precision medicine.
Studies have shown significant diagnostic errors in a range of 0.1\% up to 10\% of cases, depending on methodology and case selection \cite{diagnosticerror1, diagnosticerror2}, which are at risk to further increase with surging time pressure.

Artificial intelligence (AI) is often proposed as a solution to these challenges \cite{histo-xai-review, stenzinger2022artificial,van2021deep}. 
Seminal studies have shown that deep learning-based approaches can classify common diseases \cite{Byeon2022,Steinbuss2020,campanella2019clinical,strom2020prostate}, identify tumor origin \cite{lu2021origin}, prognosticate patient outcome \cite{courtiol2019mesothelioma,chen2022multimodal}, quantify biomarkers, and even predict certain mutations from H\&E images \cite{binder-morphological, arslan2024pancancer, kather2020pancancer}, demonstrating the great potential of AI for histopathology. 
However, all these approaches follow the paradigm of \emph{supervised learning}, that is, they require the to-be-recognized pathological patterns to be present \emph{and} labeled in the training data. As a result, existing AI solutions are limited to common diagnoses only, for which sufficiently large amounts of training examples are available.

\begin{figure*}[h!]
    \centering
    \includegraphics[width=\textwidth]{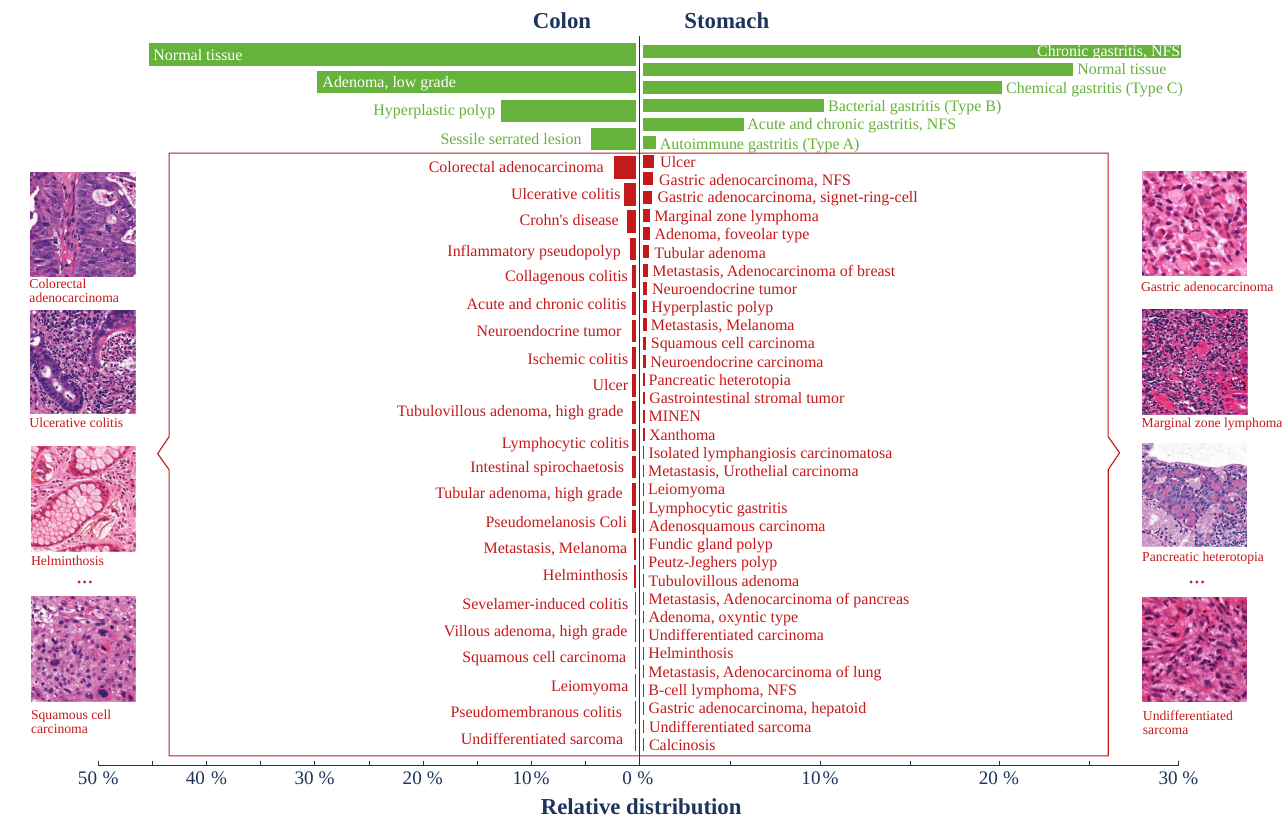}
    \caption{{\textbf{Diseases in GI biopsies.}}
   Bar plot showing the frequency distribution of diagnoses in colon and stomach biopsies in the Charité cohort. Frequent findings are highlighted in green and represent the common or ``normal'' cases (90/91\% of all cases for stomach/colon). The distribution has a long tail of infrequent/rare diagnoses or ``anomalies'' (red), which the AI-based AD approach aims to detect (NFS = not further specified, MINEN =mixed neuroendocrine-nonneuroendocrine neoplasm).}
    \label{fig:clinical-case-dist}
\end{figure*}

In clinical practice, however, only few diagnoses are common and the vast majority of diseases are relatively rare, reflected by a long-tail distribution of diseases, exemplified for colon and gastric routine biopsies in Figure \ref{fig:clinical-case-dist}. Critically, the challenge for pathologists is to reliably detect and diagnose all those infrequent diseases among the common (and therefore easy-to-diagnose) cases. Current AI models fail to support this crucial aspect, as it often is practically impossible to gather sufficient training data for the long tail of infrequent conditions. As a result, existing classifiers tend to produce false predictions on uncommon differential diagnoses or miss them completely \cite{van2021deep, evanserrorspath}. This issue is mostly ignored in the literature, where allegedly high performances are only reported on curated datasets of common findings \cite{uncertainty_kompa, uncertainty_ovadia, evanserrorspath}. 
In consequence, human confirmation is currently required for every slide subjected to AI-analysis, as less common pathologies must always be expected in routine diagnostics. We believe that this shortcoming is a major obstacle for the adoption of AI in histopathological diagnostics.

In this study, we address this fundamental problem and examine whether we can detect infrequent findings in histopathological images using AI-based anomaly detection (AD) \cite{RuffRev}.
In contrast to supervised learning, AD follows the paradigm that certain data inputs are too infrequent to be sufficiently represented during model training. Instead of trying to learn insufficiently represented patterns, AD methods aim to very precisely characterize the frequent findings, which in our setting includes normal cases and common pathologies that can be learned by supervised methods. Samples deviating from the learned common characteristics are consequently deemed ``anomalies.'' 
Since only frequent findings are used for AD model training, there is no need for extensive data collection or annotation gathering of rarer instances from the tail of the disease distribution. 

We propose different modern AD methods for histopathology and apply them to whole slide images (WSI) of gastrointestinal (GI) biopsies, which arguably pose the most frequent diagnostic question in histopathology. Here, most cases belong to one of ten common diagnoses while the remaining patients suffer from one of the many rarer diseases in the long tail (Figure \ref{fig:clinical-case-dist}), making this a particularly relevant use case. The proposed AI-AD can serve as a clinical AI assistant that flags critical cases requiring the pathologists' particular attention during routine diagnostics. Further, it may enhance the safety profile of supervised AI models and drive AI adoptation and automation.

\section{Methods}  

\subsection{Datasets}

We continuously digitized H\&E-stained slides of gastric and colon specimens from routine diagnostics at the Charité university hospital (years 2020/2021) and translated the diagnoses from the clinical reports into SNOMED-CT codes. 1973 slides with frequent findings were  included in our Charité dataset (stomach = 961, colon = 1,012) along with 200 slides presenting anomalous findings (126 from years 2020/2021, 74 from the archives of years 2000--2021). All slides were scanned with a 3DHistech P1000 scanner. Resulting dataset characteristics are presented in Appendix \ref{app:dataset_stats}. In total, 65 distinct diagnoses are represented.

We created detailed annotations for all anomalous slides for evaluation purposes. The annotations delineated (i) regions of interest containing the actual diagnosis-defining anomalies, (ii) non-diagnosis-defining anomalous regions as, e.g., tumor-adjacent inflammation, and (iii) artifacts like pen marks or blurry areas. 

Additionally, we retrieved 2,901 slides from various tissue types other than colon and stomach as auxiliary training data for the OE approach (see below). Slides were taken from routine diagnostics at the Charité pathology department, including healthy and diseased tissue from different organs (details in Appendix \ref{app:dataset_stats}).

To demonstrate the ability of our methodology to generalize across hospitals and scanners, we collected an independent dataset with 192 gastric and 157 colon slides from the Institute of Pathology, LMU Munich (more details in Appendix \ref{app:lmu_cohort}).

\subsection{Evaluation scheme and training data}
We report the mean and standard deviation from a 5-fold cross-validation evaluation. 
As cellular changes in low-grade and high-grade adenomas exist on a continuum and the detection of high-grade changes is crucial, we excluded low-grade adenomas from the training data (further details in Appendix \ref{app:multiscale}). 

All AD models operate on $340\times 340$ pixel patches from the WSI.
To aggregate the patch scores to the slide level, we selected the 10\% patches with the highest anomaly scores and computed the mean of their respective anomaly scores. 

\begin{figure*}[h!]
    \centering
    \includegraphics[width=\textwidth]{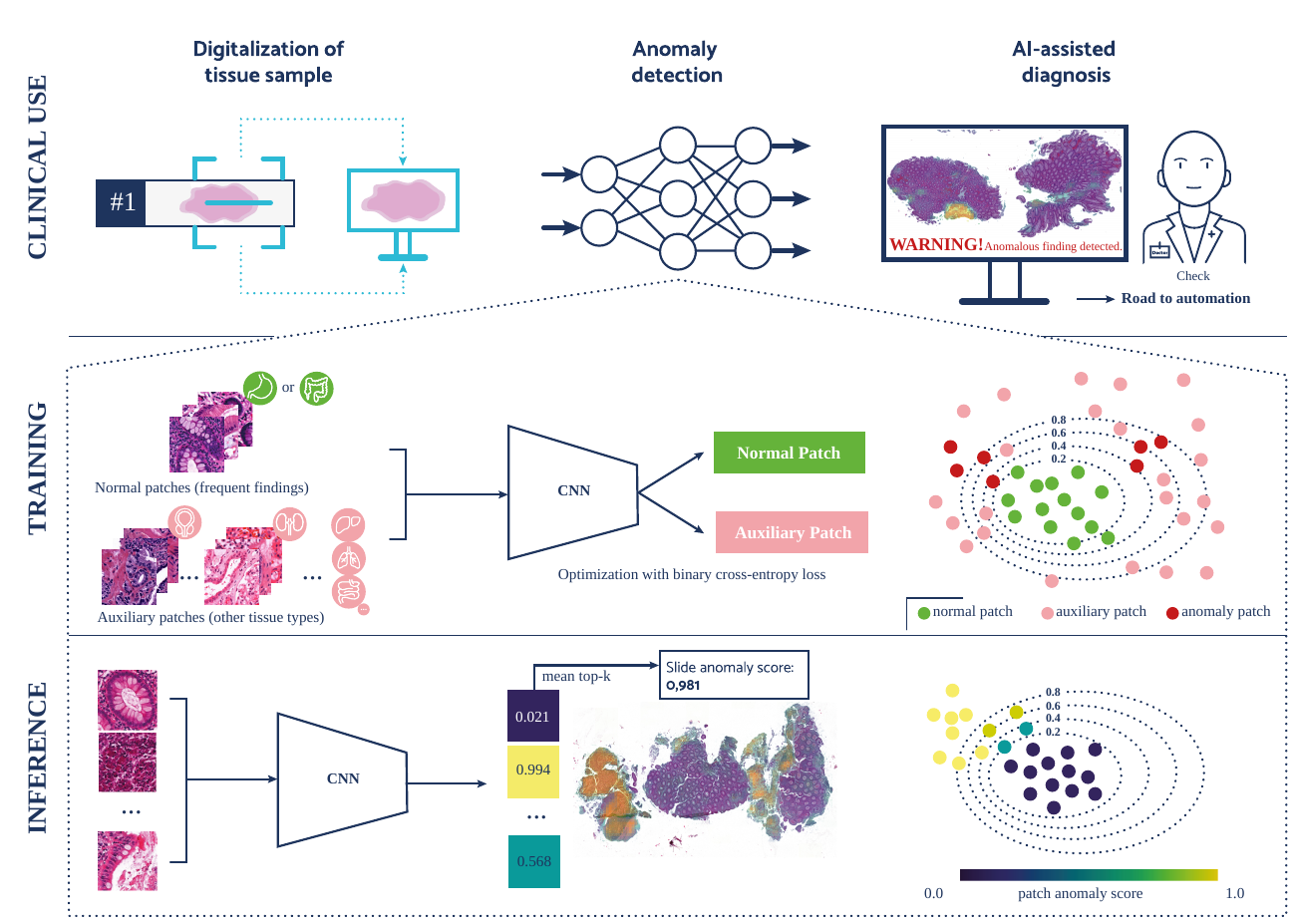}
    \caption{\textbf{Clinical use case and overview of anomaly detection approach.} \textbf{(CLINICAL USE)} Our AD approach can support pathologists in the routine diagnostic workflow by detecting abnormal cases and highlighting corresponding abnormal tissue regions. It may enable automated workflows in the future. \textbf{(TRAINING)} The OE model is trained by exposing and separating patches of frequently found (``normal'') tissue patterns in colon- and stomach biopsies from diverse patches of other tissue types (e.g., small intestine, lung, liver, prostate, breast, etc.). Through this exposure, the model learns the specific features of frequent colon and stomach findings, enclosing them into a compact decision boundary, thereby enabling the detection of anomalies as data points falling outside of this boundary. The so-trained model using ``auxiliary patches'' is subsequently able to generalize to relevant anomalous findings in colon and stomach tissue. \textbf{(INFERENCE)} During inference, we compute the anomaly score for each patch of a slide and aggregate the scores. Our approach provides the pathologist with a slide anomaly score and a heatmap of anomalous regions.}
    \label{fig:overview}
\end{figure*}

\subsection{Deep anomaly detection}

Deep AD methods learn meaningful feature maps from high-dimensional data via deep learning to distinguish ``normal'' from ``anomalous'' patterns in the learned feature space \cite{RuffRev}. Since anomalous data are naturally infrequent and hard to obtain, the models are usually trained with normal data only (unsupervised) or a few additional real anomalies (semi-supervised) \cite{RuffRev}. For this study, we restricted our methods to the unsupervised setting. We compared two deep AD paradigms: self-supervised learning and OE.

\subsubsection*{Deep AD with self-supervised learning}

The idea of self-supervised learning-based AD is to train a deep neural network on an auxiliary task such as contrasting semantically matching vs.\ different samples in feature space \cite{tack2020}. 
A trained model compresses high-dimensional images into a low-dimensional feature space, where dissimilarities can be measured and thus anomalies detected.
We assessed two variants: (i) the feature extraction model CTransPath \cite{ctranspath}, trained on 32,220 H\&E-stained diagnostic slides comprising 32 cancer subtypes from TCGA and PAIP, and (ii) the CTransPath model fine-tuned on the frequent finding patches with a deep one-class classification (OCC) loss \cite{reiss2021panda}. In each setting, we created separate models for gastric and colon data, respectively.
Further approaches, which showed inferior performance, are described in Appendix \ref{app:knn_results}. 
To determine patch anomaly scores in feature space, we applied a modified $k$-nearest neighbors (kNN) algorithm. 

\subsubsection*{Deep AD with OE}

The idea of OE is to collect vast amounts of informative auxiliary data that---unlike true anomalous data---are easy to obtain in large numbers \cite{hendrycks2019deep}. A classifier is trained to distinguish the common data from this auxiliary data. If the auxiliary data deviates from the common data distribution while still retaining a substantial degree of similarity, the model is able to learn the specific characteristics of the common class, enabling it to detect true anomalies \cite{hendrycks2019deep, liznerski2022exposing}.

To adapt OE to histopathological images, we defined all frequent finding patches of one tissue type (stomach or colon) as normal.
Patches from other collected tissue types (prostate, kidney, liver, etc.) from our separate OE dataset and normal patches from the respective other tissue type (colon or stomach) are auxiliary anomalies. 
We hypothesized that patches from a range of other tissue types are a close proxy for potential anomalies, while being sufficiently similar to the normal patches of the target tissue type. We trained a deep neural network to distinguish normal patches and these auxiliary anomalies, employing a binary cross-entropy loss \cite{liznerski2022exposing}. The model learns a compact decision boundary around patches of the frequent findings (Figure \ref{fig:overview}, TRAINING). After training, we computed the anomaly score of a patch as the probability the model assigns to the anomaly class (Figure \ref{fig:overview}, INFERENCE). We also fine-tuned the CTranspath model on the OE task, which showed similar performance as a randomly initialized ResNet-18 model (detailed ablation in Appendix \ref{app:model_variations}). 
\section{Results} 
For our study, we collected and digitized 5,423 tissue slides at two hospitals (Charité, Berlin and University Hospital of Ludwig-Maximilians-University, Munich) resulting in 17M histological images for training and evaluation purposes. 
The Charité cohort, which we used for training and primary validation, includes 2,173 GI biopsies showcasing 65 distinct diagnoses and 200 cases of anomalous findings (as specified in Figure 1). Additionally, we retrieved 2,901 slides from various tissue types other than colon and stomach as auxiliary training data, which we use in the OE approach. Dataset statistics are reported in Appendix \ref{app:dataset_stats}. 

We applied and extended modern AD methodologies \cite{RuffRev}, namely self-supervised AD \cite{tack2020} (with CTransPath \cite{ctranspath} and OCC \cite{reiss2021panda}) and OE \cite{hendrycks2019deep, liznerski2022exposing}, which have recently drastically reduced error rates on natural image benchmarks and so far had not been adopted to diagnostic pathology. Herein, we focused on the clinically relevant use case of detecting \emph{all} diagnoses in the dianostic long tail.
For training, we exclusively exposed the AD models to patches of frequent findings (as well as auxiliary slides for the OE approach). Importantly, infrequent findings (i.e., the anomalies) were not shown during training. 

AD performance results, depicting how accurately infrequent diseases were detected, are presented in Table \ref{tab:performance}. The best self-supervised method achieved slide-AUROC scores of 94.95\% (stomach) and 89.76\% (colon) and patch-AUROC scores of 89.73\% (stomach) and 87.03\% (colon). With the OE-based method, we attained higher AUROC scores than with the self-supervised methods, with slide-AUROC scores of 95.04\% (stomach) and 91.01\% (colon) and patch-AUROC scores of 91.37\% (stomach) and 90.47\% (colon).
These results demonstrate that deep anomaly detection can reliably detect long-tail diseases in histopathological slides. 

\begin{table*}[tbh]
    \footnotesize
    \caption{Performance of the three proposed anomaly detection methods on the Charité cohort. Slide-AUROC measures slide separability based on the aggregated slide anomaly scores and respective slide diagnosis labels. Patch-AUROC measures the separability of individual patches based on the patch anomaly scores and ground-truth labels provided by pathologist annotations. Reported results are mean and standard deviation based on a 5-fold cross-validation of the normal training data, expressed as percentages. }
    \label{tab:performance}
    \centering
    \begin{tabular}{llcccc}\toprule
     & &   \multicolumn{2}{c|}{\textbf{Stomach}} & \multicolumn{2}{c}{\textbf{Colon}} \\
    Model & Diagnosis Group & \multicolumn{1}{c}{slide-AUROC} & patch-AUROC & \multicolumn{1}{|c}{slide-AUROC} & patch-AUROC \\
 \midrule
Self-supervision w/ kNN &                              &        $94.95 \pm 1.16$ &        $87.21 \pm 0.36$ &        $89.76 \pm 0.77$ &        $85.09 \pm 0.63$ \\
                                       & \tiny{Neoplastic, malignant} &  \tiny{$95.23 \pm 1.0$} & \tiny{$87.03 \pm 0.47$} & \tiny{$97.48 \pm 0.47$} & \tiny{$90.99 \pm 0.64$} \\
                                       &    \tiny{Neoplastic, other} & \tiny{$94.95 \pm 1.39$} & \tiny{$92.04 \pm 0.25$} & \tiny{$95.36 \pm 0.46$} &  \tiny{$88.9 \pm 1.72$} \\
                                       &          \tiny{Inflammation} &  \tiny{$91.84 \pm 1.9$} & \tiny{$87.86 \pm 0.59$} & \tiny{$90.03 \pm 1.16$} & \tiny{$84.46 \pm 0.71$} \\
                                       &  \tiny{Other} & \tiny{$98.37 \pm 0.74$} & \tiny{$92.78 \pm 0.25$} & \tiny{$51.83 \pm 2.47$} & \tiny{$44.57 \pm 1.96$} \\
                                       \midrule
               Self-supervision w/ OCC &                              &        $93.76 \pm 1.39$ &        $89.73 \pm 0.47$ &        $88.51 \pm 0.69$ &        $87.03 \pm 0.49$ \\
                                       & \tiny{Neoplastic, malignant} & \tiny{$95.24 \pm 1.31$} & \tiny{$91.01 \pm 0.59$} &  \tiny{$96.12 \pm 0.8$} &  \tiny{$92.35 \pm 0.6$} \\
                                       &    \tiny{Neoplastic, other} & \tiny{$90.28 \pm 1.66$} & \tiny{$92.19 \pm 0.37$} & \tiny{$93.64 \pm 0.62$} & \tiny{$91.61 \pm 0.76$} \\
                                       &          \tiny{Inflammation} & \tiny{$91.11 \pm 1.55$} & \tiny{$92.96 \pm 0.53$} & \tiny{$89.94 \pm 1.04$} & \tiny{$86.73 \pm 0.51$} \\
                                       &  \tiny{Other} & \tiny{$96.95 \pm 1.07$} & \tiny{$92.17 \pm 0.51$} & \tiny{$43.93 \pm 2.16$} & \tiny{$44.85 \pm 1.52$} \\
                                       \midrule
  Outlier Exposure &                              &        $95.04 \pm 0.54$ &        $91.37 \pm 0.34$ &        $91.01 \pm 0.69$ &        $90.47 \pm 0.33$ \\
                                       & \tiny{Neoplastic, malignant} & \tiny{$97.72 \pm 0.44$} & \tiny{$95.02 \pm 0.28$} & \tiny{$96.97 \pm 0.61$} & \tiny{$96.23 \pm 0.27$} \\
                                       &    \tiny{Neoplastic, other} & \tiny{$88.45 \pm 0.82$} & \tiny{$90.51 \pm 0.48$} & \tiny{$95.72 \pm 0.91$} & \tiny{$94.17 \pm 0.38$} \\
                                       &          \tiny{Inflammation} &  \tiny{$93.4 \pm 1.02$} & \tiny{$95.75 \pm 0.34$} & \tiny{$94.42 \pm 1.07$} & \tiny{$90.24 \pm 0.41$} \\
                                       &  \tiny{Other} &  \tiny{$95.61 \pm 0.3$} & \tiny{$92.44 \pm 0.67$} & \tiny{$40.41 \pm 1.86$} & \tiny{$37.25 \pm 0.86$} \\
\bottomrule
\end{tabular}
\end{table*}

\begin{figure}[h!]
    \centering
    \textbf{\sffamily Stomach} \\
    \vspace{1mm}
    \includegraphics[width=.95\textwidth]{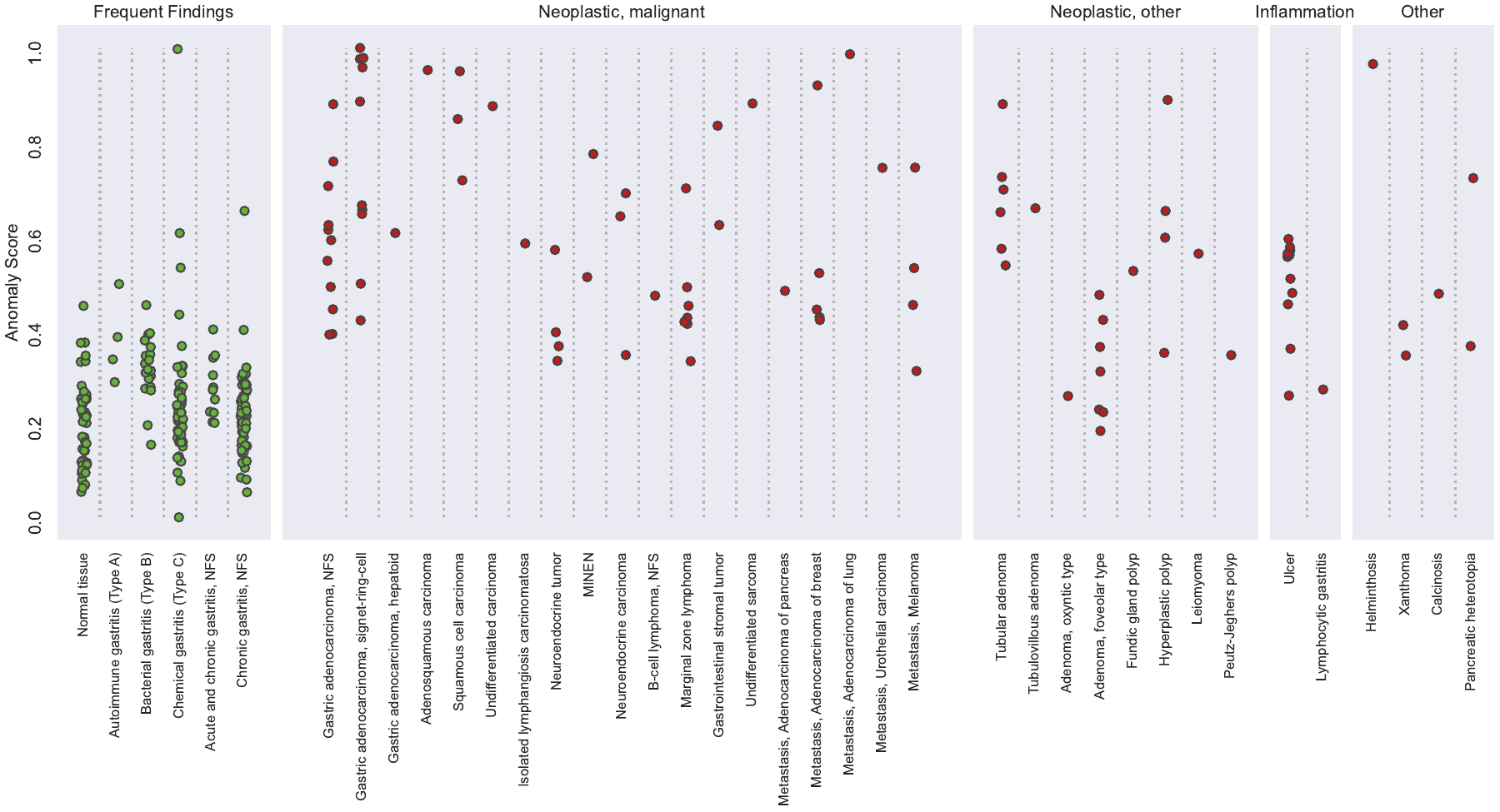}
    \vspace{1mm}
    \textbf{\sffamily Colon}
    \vspace{1mm}
    \includegraphics[width=.95\textwidth]{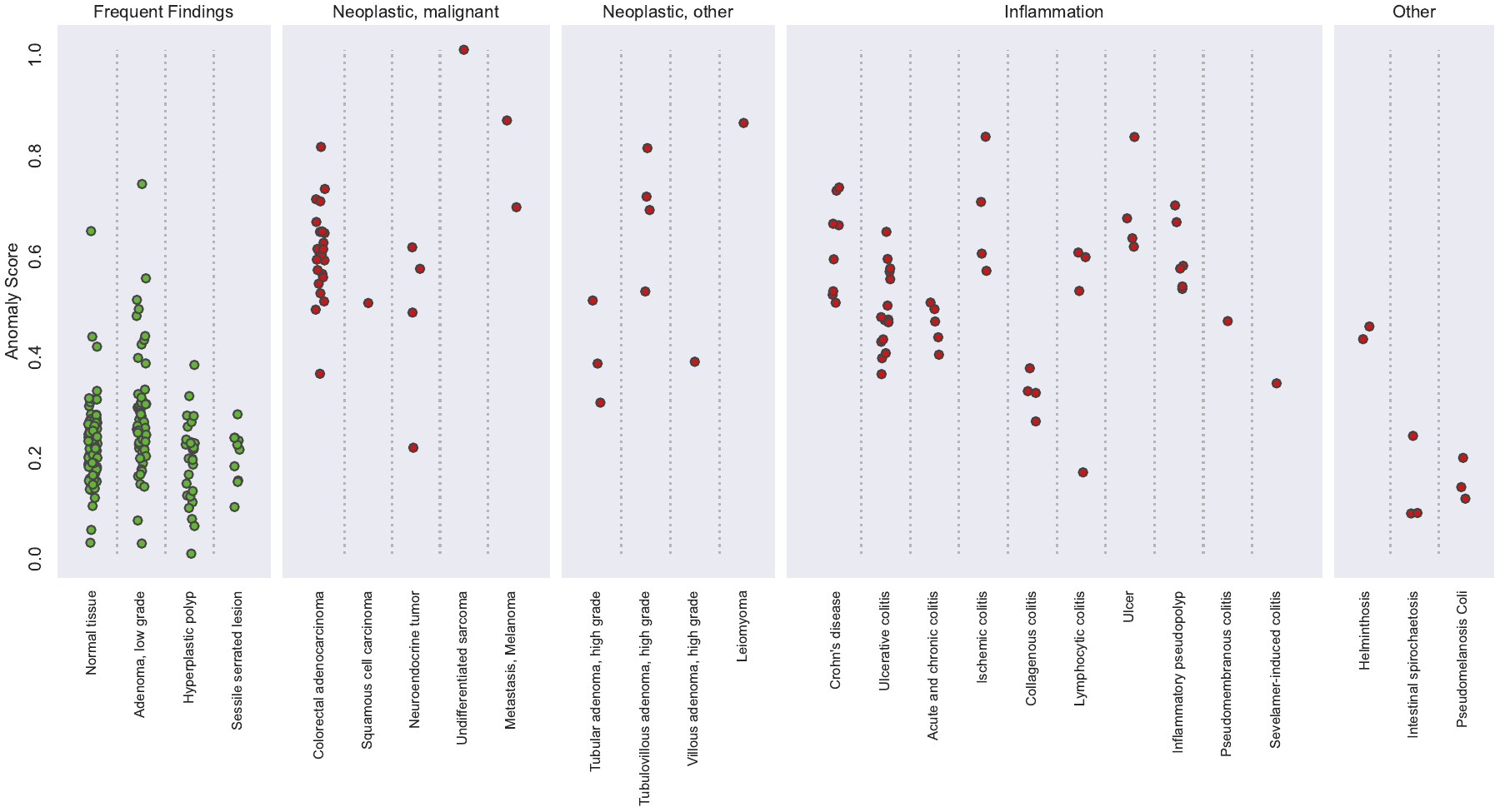}
    \caption{\textbf{Distribution of slide anomaly scores.} Slide anomaly scores for individual slides after patch aggregation, grouped by diagnoses within diagnostic groups, for the validation dataset of the Charité cohort with the OE model. Results are from one split of the 5-fold cross validation. Results for stomach biopsies are shown at the top and for colon biopsies at the bottom (NFS = not further specified, MINEN = mixed neuroendocrine-nonneuroendocrine neoplasm).}
    \label{fig:slide-dist-charite}
\end{figure}

\subsection{AI-AD detects diverse pathological patterns}
\label{pattern}

In contrast to previous work, we aimed to detect all histological tissue changes beyond common pathologies. This includes neoplasms, inflammation, infections, as well as other tissue changes likecalcinosis, xanthoma, or pancreatic heterotopia. Our approach addresses the real-world clinical setting, in which {\em any} type of pathological change has to be expected at {\em any} time.

We were largely successful in this task, as almost all diseases from various diagnostic groups resulted in considerably elevated anomaly scores (Figure \ref{fig:slide-dist-charite})
Importantly, malignant tumors of very different morphology and histogenesis, such as carcinomas, neuroendocrine tumors, lymphomas, metastatic melanomas, or sarcomas, were reliably assigned high anomaly scores. In fact, of all diagnostic groups, slide-AUROCs for malignancies were highest with $97.72 \%$ for stomach and $96.97 \% $ for colon, respectively. This is crucial, as detecting malignancy is the most consequential task in histopathological diagnostics. Infrequent benign and precancerous neoplastic changes were also reliably detected (slide-AUROC $88.45 \%$ for stomach, $95.72 \%$ for colon). Additionally, the AD model effectively recognized inflammation of the colon (slide-AUROC of $94.42$\%; for stomach most types of inflammation are frequent and therefore non-anomalous). 

Pathological alterations that were not yet recognized by the AD model are pseudomelanosis coli, a harmlessbrown discoloration of colonic mucosa, as well as intestinal spirochaetosis, a bacterial infection presenting with a thin fuzzy line of bacteria on the surface of colon epithelium. 

Importantly, for the given results on the Charité cohort, setting a conservative detection threshold on the slide anomaly scores to achieve 100\% (99\%, 95\%) anomaly sensitivity, already 36.2\% (51.91\%, 72.93\%) of stomach and 4.21\% (4.99\%, 38.2\%) of colon cases can be reliably predicted as not being abnormal.
Excluding anomalous cases of pseudomelanosis coli and intestinal spirochaetosis, which are either of no clinical significance (pseudomelanosis) or usually require special stains (spirochaetosis) even for trained pathologists, 
the colon numbers increase to 22.29\% (49.46\%, 85.22\%). These results demonstrate the potential of the presented AD methodology for pathologist time savings and safe automation in diagnostics.

\subsection{Heatmaps enable visual feedback and interpretation of AI-AD}
Interpretability of AI predictions is crucial in a medical setting \cite{histo-xai-review, bach2015pixel, samek2021explaining}, and it is important to guide experts to the anomalous patterns. Furthermore, it is important to verify that the model's predictions are not based on shortcut features, e.g. tissue artifacts (so-called ``Clever Hans'' effects \cite{lapuschkin2019unmasking, kauffmann2020clever}).

\begin{figure}[H]
    \centering
    \includegraphics[width=0.93\textwidth]{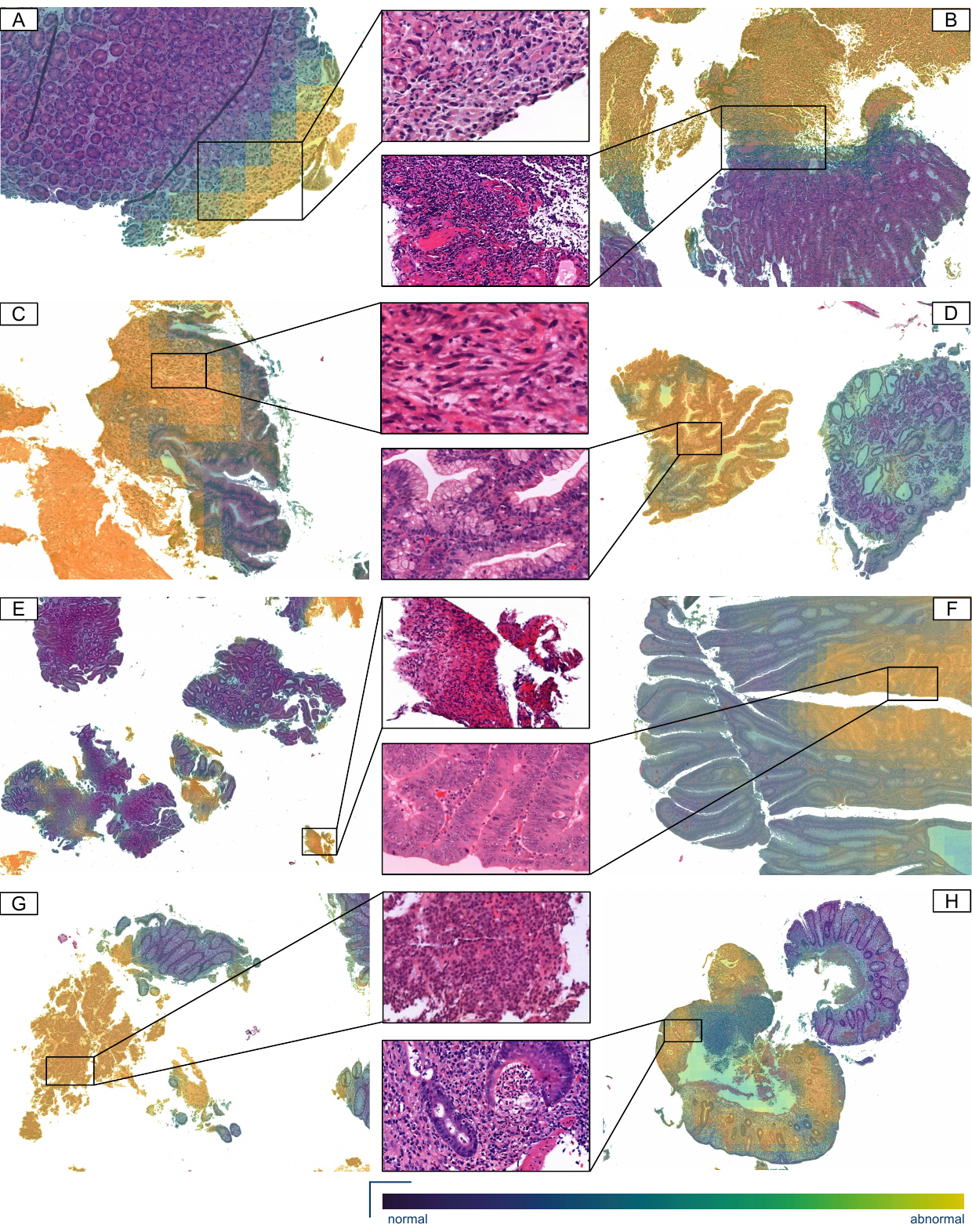}
    \caption{\textbf{Heatmap visualization of AD results} \textbf {(a-c)} Morphological distinct correlates of adenocarcinoma (a),  marginal zone lymphoma (b), and undifferentiated sarcoma (c) in stomach tissue are highlighted as abnormal. Common tissue artifacts in (a) (tissue folds showing as dark vertical lines) do not considerably influence the heatmap \textbf{(d-e)} Benign changes of gastric foveolar adenoma (d) and changes consistent with a stomach ulcur (e) are highlighted as abnormal.  \textbf{(f)} High-grade epithelial dysplasia of colon tissue is differentiated from adjacent low-grade dysplasia and highlighted as abnormal, showcasing the AD's ability to differentiate between subtle histological variations.  \textbf{(g)} A tissue fragment of a neuroendocrine tumor of the colon is highlighted as abnormal. \textbf{(h)} Colon tissue with inflammation is highlighted as abnormal.}
    \label{fig:heatmaps}
\end{figure}

Our patch-based approach allowed us to create heatmaps that highlight regions determined anomalous by our model. Exemplary heatmaps, which showcase the reliable detection performances for a broad range of anomalous malignant and benign findings, are shown in Figure \ref{fig:heatmaps}. Additional heatmaps of complete tissue cuts along with pathologists' annotations are provided in Appendix \ref{app:add_heatmaps}.

Interestingly, tissue artifacts were often not or only partly highlighted in heatmaps, which is in line with a lower patch-AUROC for artifacts than for anomalies (Appendix \ref{app:artifacts}). This is important, as such artifacts can be common among slides and should therefore not result in markedly enhanced slide anomaly scores or uninformative heatmaps.

\subsection{AI-AD generalizes across hospitals and scanners}
Research indicates that the performance of AI models can drastically deteriorate with changes in input data characteristics \cite{hendrycks2021many}. In histopathology, variations due to different lab staining protocols or differences in scanner equipment are known critical factors \cite{campanella2019clinical}.
To test the generalization performance of our trained models, we evaluated their performance on an independent cohort from the Institute of Pathology at Ludwig-Maximilians-University (LMU) Munich, where slides were digitized with a different scanner type. 
We collected cases with frequent findings (63 for colon, 164 for stomach) as well as cases presenting anomalous findings (94 for colon, 28 for stomach), resulting in a total of more than 500k image patches for generalization validation purposes (Appendix \ref{app:dataset_stats}).

Evaluating model performance--\emph{without re-training the models on the new data}--yielded competitive slide-AUROCs of 94.5\% (stomach) and 85.88\% (colon) for our previously best-performing approach (detailed results in Appendix \ref{app:lmu_results}).
Focusing the validation analysis on the clinically most relevant anomalies, i. e. malignant pathologies, we reach slide-AUROCs of $94.77 \%$ for stomach and $95.02 \%$ for colon.

\section{Discussion} 
The long-tail disease distribution encountered in clinical reality (only few diseases are common, most diseases are less common) poses a significant challenge for the implementation of AI in medical diagnostics. It is often impossible to accurately represent all diseases in the long diagnostic tail during model training and, in most cases, not even attempted. This causes critical diagnostic errors by AI models, compromising security and clinical usability \cite{van2021deep, evanserrorspath}. Our developed AD for histopathology addresses this critical shortcoming, as it does not depend on training data for the long-tail diseases. We were able to show high detection performances within GI biopsies and demonstrated generalization across labs, staining patterns, and scanner characteristics.

First promising adaptations of AD have been presented in dermatology and radiology \cite{dermatology,radiology}. However, AD has been largely unexplored for histopathology so far. In previous publications, the focus has been on the detection of single pathologies defined as anomalous, as for example breast cancer metastases in lymph nodes \cite{pocevivciute2021unsupervised,shvetsova2021anomaly,stepec2021unsupervised, linmans2024diffusion}. In contrast, our dataset with a large number of different infrequent findings reflects the clinical reality, and thereby stands out from the related AD work in histopathology.
Additionally, there are methodological limitations, as previous work is predominantly not based on the recent advances of OE or self-supervised learning, with the exception of \cite{ZINGMANpublished} who consider a variant of the latter for the detection of non-alcoholic fatty liver disease in mouse liver tissue. Rather, most previous works use AD models such as generative adversarial networks \cite{pocevivciute2021unsupervised, stepec2021unsupervised,zehnder2022multiscale}, autoencoders \cite{shvetsova2021anomaly}, or flow-based models \cite{pawlowski2021abnormality}, which have been found less effective \cite{RuffRev}. 

Our results further provide interesting insights into state-of-the-art AD methods on histopathological data: (1) we show that the right trade-off between similarity and diversity of the OE data is crucial for generalization, (2) only about 100 slides of common findings are sufficient for characterization of the normal training data, and (3) color augmentations and stain normalization are critical for generalization to different scanners and hospitals. We present respective ablation studies for these insights in Appendix \ref{app:ablations}.

There are certain limitations to our AI-AD approach, particularly with respect to the detection of extremely subtle tissue changes in colon biopsies. These include pseudomelanosis coli, collagenous/lymphocytic colitis, and intestinal spirochaetosis, which are difficult to detect even for the trained expert and, in the case of collagenous/lymphocytic colitis and spirochaetosis, require additional special stains. While lower anomaly scores are therefore not surprising in such cases, additional strategies will need to be developed in the future to improve model performance further. The implementation of semi-supervised learning methodologies may hold significant potential in this aspect \cite{ruff2020dsad}. As some pathologies, such as architectural changes in colon mucosa,  are hardly detectable on individual patches of the currently used size -- even for pathologists -- multi-scale approaches with larger tissue context should be explored.

Our AD model used as a stand-alone clinical AI assistant has the potential to substantially improve both diagnostic efficiency and quality by reducing the amount of missed diagnoses through identifying ``suspicious'' cases and highlighting anomalies in histological slides. Critically, because of its design, it can be expected to reliably detect \emph{any} kind of primary or metastatic cancer in stomach/colon samples even beyond the entities we evaluated it on. To our knowledge, no other published AI tool is capable of this in a zero-shot manner, even across other tissues.
An integrated approach of AD and supervised detection of common findings (for GI-samples e.g. \cite{Byeon2022, Steinbuss2020}) could in the future enhance the safety profile of supervised models and even lead to an overall safe automatic processing of samples. Our results indicate that with the current performance, already up to a third of biopsies with frequent findings could be automatically diagnosed without the risk of missing any less frequent and potentially severe diseases. This fraction can be expected to grow with future model improvements and may ultimately only leave a subset of cases to require manual review, which could drastically reduce pathologists' workload and pave the way for a largely automated and safe AI-based histopathological diagnostics.

\paragraph{Acknowledgements}
This work was partly funded by the German Ministry for Education and Research (under refs 01IS14013A-E, 01GQ1115, 01GQ0850, 01IS18056A, 01IS18025A and 01IS1-8037A) and BBDC/BZML and BIFOLD. 
NP is participant in the BIH Charité Junior Digital Clinician Scientist Program funded by the Charité – Universitätsmedizin Berlin, and the Berlin Institute of Health at Charité (BIH).
KRM was partly supported by the Institute of Information \& Communications Technology Planning \& Evaluation (IITP) grants funded by the Korea Government (MSIT) (No. 2019-0-00079, Artificial Intelligence Graduate School Program, Korea University and No. 2022-0-00984, Development of Artificial Intelligence Technology for Personalized Plug-and-Play Explanation and Verification of Explanation). 
Part of this work was conducted within the DFG research unit FOR 5359 (KL 2698/6-1 and KL 2698/7-1). MK acknowledges support by the Carl-Zeiss Foundation, the DFG awards KL 2698/2-1 and KL 2698/5-1, and the BMBF awards 03|B0770E and 01|S21010C.

\paragraph{Ethics approval}
The current project has been considered by the local ethics committees in Berlin and Munich under the EA1/125/23 and 23-0334, respectively.

\paragraph{Author contributions}
J.D., N.P., P.L., L.R., K.-R.M., F.K. conceptualized the project. N.P, F.K, O.B., L.R. were responsible for the data collection. N.P., F.K. performed data curation, data annotation and data analysis. J.D., J.H., P.L. carried out data analysis and model construction; J.D., J.H, N.P., P.L. performed model validation and data visualization. T.W. helped with dataset building. N.P., J.D., J.H., P.L., L.R. K-R.M., F.K. wrote the first draft, which was reviewed and edited by T.W., S.S., M.K., O.B., D.H., M.A.; K.-R.M., F.K. and L.R. supervised the project. Funding was secured by N.P., K.-R.M., M.K., and F.K.
Correspondence to L.R., K.-R.M., F.K.

\printbibliography

\newpage
\appendix

\section{Related work} \label{app:related_work}
This section reviews related work in more detail and provides a thorough introduction to the latest deep anomaly detection methods and their applications in histopathology. 

\subsection{Deep Anomaly Detection (AD)} \label{app:related_work_ad}
In contrast to classical approaches to AD that are known to perform poorly on high-dimensional data such as images \cite{RuffRev}, modern AD methods employ deep neural networks that scale well with higher dimensions \cite{ruff2018deep,pang2021,RuffRev}. 
The first deep methods used autoencoders \cite{sakurada2014anomaly,zhou2017anomaly,nguyen2019}, generative models \cite{schlegl2017unsupervised,deecke2018image,schlegl2019}, and one-class models \cite{ruff2018deep,wu2019deep} to perform AD. 
Autoencoders are neural networks trained to compress inputs into a low-dimensional space, from which they then reconstruct the original input.
Trained only on normal samples, autoencoders are less able to accurately reconstruct anomalous patterns. The difference between the original input and the reconstruction can be used for determining an anomaly score. 
Generative models can be used for AD in a similar manner; e.g. by taking the difference of the most similar generated sample to the original. 
In deep one-class models, a neural network is trained to map normal samples to a latent space so that the samples are encompassed in a hypersphere of minimal radius. For inference, samples that lie outside of this hypersphere are considered anomalous. 
Recently, self-supervised learning \cite{golan2018deep,hendrycks2019using,tack2020} and outlier exposure \cite{hendrycks2019deep,liznerski2022exposing} caused a breakthrough in deep AD, halving the error rates on established benchmarks \cite{tack2020, liznerski2022exposing}. 

\paragraph{Self-supervised Anomaly Detection}
Self-supervised learning emerged as a means to learn general-purpose features for various tasks without the need for manual labeling.
In self-supervised learning, labels for learning are automatically generated.
For example, in \cite{gidaris2018unsupervised}, the authors rotated each training image with four different angles and trained the neural network to predict the angle.
They termed the trained network ``RotNet.'' 
RotNet is also the first method that was used for deep AD \cite{ golan2018deep}, marking the beginning of self-supervised deep AD. 
Training RotNet exclusively on normal samples leads to uncertainy when applied to anomalies. 
Thus, the uncertainty of the prediction can be used as an anomaly score. 
Later works improved upon this by adding and combining additional geometric transformations \cite{hendrycks2019using}.
More recent approaches used contrastive losses in combination with image transformations to further improve the AD performance \cite{tack2020, sohn2021, zou2022spot}.
In the CSI method \cite{tack2020} the contrastive loss of SimCLR is used \cite{chen2020simple} to train a network to map diverse transformations of normal images close together, while pushing different normal samples apart.
The network also has to predict the kind of transformation, as was done in \cite{hendrycks2019using}, and integrates a k-nearest neighbor algorithm for final detection.
CSI is still the state-of-the-art AD method for natural images, when one is restrained from using extra training data in the form of outlier exposure or pretraining.

\paragraph{Outlier Exposure}
AD algorithms are typically unsupervised, because it is considered unfeasible to find anomalous data that sufficiently represents \emph{everything} anomalous. 
An anomaly detector trained with samples that represent just a subset of all notions of anomalousness is prone to be biased towards the seen notions, generalizes poorly, and thus performs inaccurately overall \cite{ruff2020dsad}.
However, \cite{hendrycks2019deep} found that random data samples from a domain (e.g. random images from the web for natural image tasks) are most likely anomalous for any given AD problem in the domain. 
Using a huge corpus of such ``auxiliary anomalies'' during training, leads to effective generalization and substantially enhances the performance of anomaly detectors on natural images benchmarks.
They called their approach ``outlier exposure" (OE).
Later works investigated the behavior of models trained with OE \cite{liznerski2022exposing} and found that training a binary classifier with a standard cross-entropy loss to distinguish between normal data and OE samples yields the best results. 
They also proposed a modification of the unsupervised deep one-class loss \cite{ruff2018deep}, which they termed ``hypersphere classification'' (HSC), that performs slightly worse, but is more robust to non-representative OE sample distributions. 
This is particularly important when only a few OE samples are available.
While OE with random samples from the internet is the state of the art for natural image AD, it is less suitable for subtle anomalies such as defective versions of normal samples (e.g., a cracked screw for screws being normal). 
Here, random images from the internet are too dissimilar from the data and thus too easy to detect as anomalous. 
It was found that the most effective kind of OE for this setup can be generated synthetically by perturbing normal samples to look like defective versions
\cite{goyal2020drocc,liznerski2021explainable,mirzaei2022fake}. The powerful idea of outlier exposure has not yet been explored in histopathology.

\subsection{Explaining deep AD with heatmaps} \label{app:related_work_xai}
In our paper, we provide heatmaps that represent the anomaly scores for overlapping patches, thereby providing an interpretable AD method.
Explaining deep anomaly detectors via heatmaps is a rather recent line of research. It draws from the broad literature on explainable AI (see e.g.~\cite{baehrens2010explain,simonyan2013deep,bach2015pixel,sundararajan2017axiomatic,lapuschkin2019unmasking,gunning2019xai,samek2021explaining,holzinger2022xxai,liznerski2024reimagining,varshneya2024interpretable}).
The most common approach is to make the neural network directly attribute features \cite{liznerski2021explainable}, so that each feature (e.g., pixel) is assigned a separate anomaly score, together forming the anomaly heatmap. 
Early works used generative models or autoencoders, where the pixel-wise reconstruction error yields the heatmap \cite{baur2018deep, bergmann2019mvtec, dehaene2020Iterative}. 
In a more modern approach, \cite{kauffmann2020towards,liznerski2021explainable} utilized the properties of fully convolutional neural networks to have the network directly output pixel-wise anomaly scores. 
Other works used gradient-based explanations (e.g., Grad-CAM \cite{selvaraju2017grad}) to highlight the regions the AD model focuses on, usually aligning with the more anomalous regions \cite{venkataramanan2020attention,li2021cutpaste}.
In combination with synthetically generated anomalies that are perturbed versions of normal samples, one can also directly train a semantic segmentation model that assigns a probability of anomalousness per pixel \cite{schluter2022natural}.
The training target is the pixel-wise difference between the generated anomaly and the normal sample.
In the most recent approaches, an arbitrary feature learner is trained, and then the discrepancy between extracted feature maps for a test sample and a collection of normal feature maps is used to assign an anomaly score per pixel \cite{defard2021padim,roth2022towards,zou2022spot}.

\subsection{AD in histopathology} \label{app:related_work_histo}

Until recently, applying AD to histopathology remained largely unexplored. 
Virtually all previous publications focused on detecting single disease entities as anomalous \cite{pocevivciute2021unsupervised,shvetsova2021anomaly,stepec2021unsupervised,he2022review}, such as the detection of breast cancer metastasis within lymph nodes. 
For this narrow task, an AUROC of up to 94.7\% was reached \cite{stepec2021unsupervised}, using image-to-image translation. 
Zehnder et al. aimed to detect three different kinds of anomalous changes (necrosis, peritonitis, inflammation) in mouse liver tissue \cite{zehnder2022multiscale}. 
However, the size of the used dataset was limited (total size of 50 WSI's) and included down to only 53 tiles for individual abnormal classes. 
In this regard, our diverse stomach \& colon datasets with 1.973 WSI's of frequent findings and 200 WSI's of anomalous findings provide a much larger data basis.
Methodologically, previous publications focused on AD models such as generative adversarial networks \cite{vstepec2020image,pocevivciute2021unsupervised, stepec2021unsupervised, zehnder2022multiscale}, autoencoders \cite{shvetsova2021anomaly}, or flow-based models \cite{pawlowski2021abnormality}. 
Zingman et al. have used AD to detect anomalous liver tissue showing patterns of non-alcoholic fatty liver disease in mice \cite{ZINGMANpublished}. 
Interestingly, they also used other tissue types (liver, brain, kidney, heart, lung, pancreas, spleen) during the training process of their model. However, their approach was to use other tissue types to optimize image representations with subsequent AD using a one-class classifier. 
This differs significantly from our outlier exposure approach.

If a model that detects anomalies also differentiates between different classes of normal inputs, it is typically called an out-of-distribution (OOD) detector. 
In contrast to AD models, OOD detectors usually require labels and use the confidence of the classifier to detect OOD samples.
Linsmans et al. trained a large set of models to differentiate between normal lymph node tissue and breast cancer metastasis. 
They then analyzed the models ability to accurately detect OOD-inputs of diffuse large B-cell lymphoma.
The best-performing model, a 5-multi-head ensemble, reached an AUROC of 81.02\% \cite{linmans2023predictive}. 
Dolezal et al. tested the ability to detect OOD inputs for a classifier trained to differentiate between lung squamous cell carcinoma and lung adenocarcinoma.
Interestingly, even though their model was implemented to predict the uncertainty of the classifier accurately, 21.5 \% of input slides showing non-lung, non-adenocarcinoma, non-squamous OOD cancer types were incorrectly assigned to in-distribution classes with high confidence \cite{uncertainty_dolezal}.

\section{Additional details} \label{app:details}

\subsection{Dataset statistics} \label{app:dataset_stats}
Table \ref{tab:charite-slide-dist} contains an overview over the Charité cohort including auxiliary slides used for our outlier exposure model. For the LMU cohort, Table \ref{tab:lmu-slide-dist} provides further diagnosis statistics.

\begin{table}[h!]
\caption{Diagnosis and tissue distribution in the Charité cohort.}
\footnotesize
\label{tab:charite-slide-dist}
\centering

\begin{minipage}[t]{0.48\textwidth}
\subfloat[Stomach]{%
\hspace{2em}
\begin{tabular}{lr}\toprule
\textbf{Diagnosis} & \textbf{Slides} \\
\midrule
Frequent findings & 961 \\
\hspace{3mm} Chronic gastritis, NFS & 322 \\
\hspace{3mm} Normal tissue & 230 \\
\hspace{3mm} Chemical gastritis (Type C) & 213 \\
\hspace{3mm} Bacterial gastritis (Type B) & 113 \\
\hspace{3mm} Acute and chronic gastritis, NFS & 65 \\
\hspace{3mm} Autoimmune gastritis (Type A) & 18 \\
\midrule
Neoplastic, malignant & 60 \\
\hspace{3mm} Gastric adenocarcinoma, NFS & 11 \\
\hspace{3mm} Gastric adenocarcinoma, signet-ring-cell & 10 \\
\hspace{3mm} Marginal zone lymphoma & 7 \\
\hspace{3mm} Metastasis, Adenocarcinoma of breast & 5 \\
\hspace{3mm} Neuroendocrine tumor & 4 \\
\hspace{3mm} Metastasis, Melanoma & 4 \\
\hspace{3mm} Squamous cell carcinoma & 3 \\
\hspace{3mm} Neuroendocrine carcinoma & 3 \\
\hspace{3mm} Gastrointestinal stromal tumor & 2 \\
\hspace{3mm} MINEN & 2 \\
\hspace{3mm} Isolated lymphangiosis carcinomatosa & 1 \\
\hspace{3mm} Gastric adenocarcinoma, hepatoid & 1 \\
\hspace{3mm} B-cell lymphoma & 1 \\
\hspace{3mm} Undifferentiated sarcoma & 1 \\
\hspace{3mm} Undifferentiated carcinoma & 1 \\
\hspace{3mm} Metastasis, Adenocarcinoma of lung & 1 \\
\hspace{3mm} Adenosquamous carcinoma & 1 \\
\hspace{3mm} Metastasis, Urothelial carcinoma & 1 \\
\hspace{3mm} Metastasis, Adenocarcinoma of pancreas & 1 \\
\midrule
Neoplastic, other & 22 \\
\hspace{3mm} Adenoma, foveolar type & 7 \\
\hspace{3mm} Tubular adenoma & 6 \\
\hspace{3mm} Hyperplastic polyp & 4 \\
\hspace{3mm} Adenoma, oxyntic type & 1 \\
\hspace{3mm} Tubulovillous adenoma & 1 \\
\hspace{3mm} Peutz-Jeghers polyp & 1 \\
\hspace{3mm} Leiomyoma & 1 \\
\hspace{3mm} Fundic gland polyp & 1 \\
\midrule
Inflammation & 12 \\
\hspace{3mm} Ulcer & 11 \\
\hspace{3mm} Lymphocytic gastritis & 1 \\
\midrule
Other & 6 \\
\hspace{3mm} Pancreatic heterotopia & 2 \\
\hspace{3mm} Xanthoma & 2 \\
\hspace{3mm} Helminthosis & 1 \\
\hspace{3mm} Calcinosis & 1 \\
\bottomrule
\end{tabular}}
\end{minipage}
\hspace{5mm}
\begin{minipage}[t]{0.48\textwidth}
\subfloat[Colon]{%
\begin{tabular}{lr}\toprule
\textbf{Diagnosis} & \textbf{Slides} \\
\midrule
Frequent findings & 1012 \\
\hspace{3mm} Normal tissue & 507 \\
\hspace{3mm} Adenoma, low grade & 321 \\
\hspace{3mm} Hyperplastic polyp & 139 \\
\hspace{3mm} Sessile serrated lesion & 45 \\
\midrule
Neoplastic, malignant & 31 \\
\hspace{3mm} Colorectal adenocarcinoma & 23 \\
\hspace{3mm} Neuroendocrine tumor & 4 \\
\hspace{3mm} Metastasis, Melanoma & 2 \\
\hspace{3mm} Squamous cell carcinoma & 1 \\
\hspace{3mm} Undifferentiated sarcoma & 1 \\
\midrule
Neoplastic, other & 9 \\
\hspace{3mm} Tubulovillous adenoma, high grade & 4 \\
\hspace{3mm} Tubular adenoma, high grade & 3 \\
\hspace{3mm} Leiomyoma & 1 \\
\hspace{3mm} Villous adenoma, high grade & 1 \\
\midrule
Inflammation & 52 \\
\hspace{3mm} Ulcerative colitis & 15 \\
\hspace{3mm} Crohn's disease & 8 \\
\hspace{3mm} Inflammatory pseudopolyp & 6 \\
\hspace{3mm} Acute and chronic colitis & 5 \\
\hspace{3mm} Lymphocytic colitis & 4 \\
\hspace{3mm} Ischemic colitis & 4 \\
\hspace{3mm} Collagenous colitis & 4 \\
\hspace{3mm} Ulcer & 4 \\
\hspace{3mm} Sevelamer-induced colitis & 1 \\
\hspace{3mm} Pseudomembranous colitis & 1 \\
\midrule
Other & 8 \\
\hspace{3mm} Pseudomelanosis Coli & 3 \\
\hspace{3mm} Intestinal spirochaetosis & 3 \\
\hspace{3mm} Helminthosis & 2 \\
\bottomrule
\end{tabular}}
\vspace{4mm}
\subfloat[Outlier Exposure slides]{%
\begin{tabular}{lrr}
    \toprule
    \textbf{Category} & \textbf{Slides} \\
    \midrule
    Auxiliary (for ``Outlier Exposure'') & 2,901 \\
    \hspace{2mm} Prostate & 599 \\
    \hspace{2mm} Kidney & 459 \\
    \hspace{2mm} Liver & 402 \\
    \hspace{2mm} Lung & 401\\
    \hspace{2mm} Small intestine & 283 \\
    \hspace{2mm} Breast & 204 \\
    \hspace{2mm} Other, mixed & 553 \\
    \bottomrule
\end{tabular}}
\end{minipage}

\end{table}

\begin{table}[h!]
\centering
\footnotesize
\caption{Diagnosis distribution of the colon (left) and stomach (right) LMU cohort.}
\label{tab:lmu-slide-dist}
\begin{tabular}{lr}\toprule
\textbf{Diagnosis} & \textbf{Slides} \\
\toprule
Frequent findings & 63 \\
\hspace{3mm} Adenoma, low grade & 35 \\
\hspace{3mm} Hyperplastic polyp & 13 \\
\hspace{3mm} Normal tissue & 11 \\
\hspace{3mm} Sessile serrated lesion & 4 \\
\midrule
Neoplastic, malignant & 30 \\
\hspace{3mm} Colorectal adenocarcinoma & 25 \\
\hspace{3mm} Neuroendocrine tumor & 3 \\
\hspace{3mm} Metastasis, Adenocarcinoma of lung & 1 \\
\hspace{3mm} Metastasis, Melanoma & 1 \\
\midrule
Neoplastic, other & 6 \\
\hspace{3mm} Tubulovillous adenoma, high grade & 6 \\
\midrule
Inflammation & 55 \\
\hspace{3mm} Ulcerative colitis & 23 \\
\hspace{3mm} Crohn's disease & 17 \\
\hspace{3mm} Collagenous colitis & 12 \\
\hspace{3mm} Chronic colitis, NFS & 2 \\
\hspace{3mm} Pseudomembranous colitis & 1 \\
\midrule
Other & 3 \\
\hspace{3mm} Pseudomelanosis Coli & 3 \\
\bottomrule
\end{tabular}
\hspace{2em}
\begin{tabular}{lr}\toprule
\textbf{Diagnosis}  &  \textbf{Slides} \\
\toprule
Frequent findings & 164 \\
\hspace{3mm} Chronic gastritis, NFS & 65 \\
\hspace{3mm} Normal tissue & 53 \\
\hspace{3mm} Bacterial gastritis (Type B) & 34 \\
\hspace{3mm} Chemical gastritis (Type C) & 12 \\
\midrule
Neoplastic, malignant & 24 \\
\hspace{3mm} Marginal zone lymphoma & 8 \\
\hspace{3mm} Gastric adenocarcinoma, signet-ring-cell & 7 \\
\hspace{3mm} Gastric adenocarcinoma, NFS & 5 \\
\hspace{3mm} Neuroendocrine tumor & 3 \\
\hspace{3mm} Metastasis, Adenocarcinoma of ovary & 1 \\
\midrule
Neoplastic, other & 2 \\
\hspace{3mm} Hyperplastic polyp & 2 \\
\midrule
Inflammation & 2 \\
\hspace{3mm} Ulcer & 2 \\
\bottomrule
\vspace{6em}
\end{tabular}
\end{table}

\subsection{Training details} \label{app:training_details}


To fine-tune self-supervised learning models with a one-class, we followed the training procedure of PANDA \cite{reiss2021panda}. We used a learning rate of $10^{-2}$, a batch size of 32, and the SGD optimizer. To prevent a collapse of the representations, we clipped the gradient norm to $10^{-3}$, froze the first blocks of the network, and did not update batch norm statistics during fine-tuning.


For OE, we trained the network with the standard SGD optimizer with momentum $0.9$, used a learning rate of $5 \cdot 10^{-4}$, batch size of 32, and a weight decay of $0.0001$. For augmentation, we used resized crops and color jitter. To achieve a balance between patches of frequent GI findings and OE data, we sampled an equal number of both in each batch. The OE patches were further sampled to include 50\% of near-tissue types (stomach, colon, small intestine) and 50\% of far-tissue types (all other tissues), which we defined according to their informativeness of morphological similarity. As some basic tissues (e.g. connective tissue, muscle tissue) can be found in both our frequent findings and the OE data, we aimed to remove patches with such overlapping tissue components. We did so by computing the similarity between patches from both groups and consequently removed samples from the OE data with a cosine similarity of more than $0.9$. Each model was trained within a single day on an A100 NVIDIA GPU.

\subsection{Evaluation Details} \label{app:evaluation_details}
To obtain robust patch-level results, we applied test time augmentation as an ensembling technique \cite{shorten2019survey}. We used the same augmentations as during training (random crops, color jitter) to generate $n=10$ views of the same patch, computed the anomaly score for each view, and then averaged over all the views for a final patch anomaly score.

To point pathologists at anomalous regions within a slide, we generated fine-grained anomaly score heatmaps from the patch predictions. First, we extracted patches from WSIs with an overlap of 75 pixels. Second, the patches were pre-processed and passed through the model with test-time augmentation as described above. The resulting patch scores were then aggregated into a spatial map, where the scores of overlapping patches were averaged, creating a smooth heatmap. A color was assigned to each tissue patch based on the anomaly score.

\subsection{Image pre-processing pipeline} \label{app:preprocessing}

For each collected slide, we computed a tissue boundary using standard computer vision operations. Subsequently, we extracted patches of size $340\times 340$ pixels from the identified tissue regions at 20x magnification, corresponding to a resolution of roughly 0.5 microns per pixel (mpp). We ignored patches with more than 80\% background and applied Reinhard's stain normalization method \cite{reinhard2001stainnorm} to each patch with the average stain statistics of our frequent findings as a normalization target.

\subsection{Independent validation on the LMU Munich cohort} \label{app:lmu_cohort}

To demonstrate the ability of our methodology to generalize across hospitals and scanners, we collected an independent dataset with H\&E-stained slides of 192 gastric and 157 colon specimens from the archives of the Institute of Pathology, LMU Munich, from routine diagnostics between years 2020 and 2023. Details of the LMU dataset are provided in Table \ref{tab:lmu-slide-dist}.

The LMU slides were scanned with a different scanner (Leica Aperio GT 450) and, after digitization, pre-processed in the same manner as described above. Slides with pronounced lab-specific tearing artifacts that were markedly distinct from those encountered in the Charité cohort, were excluded (for both frequent and anomalous findings). Inflammatory colon changes were graded on a scale of 0 (no inflammation) to 3 (high inflammation) and matched to inflammation levels in the Charité cohort in order to increase comparability.
Consequently, we applied our trained AD models to all patches from the slides and aggregated the patch scores to slide scores using the same strategy as on the Charité cohort. No re-training was performed; that is, no model was exposed to any of the LMU data at training time. For each assessed method, we evaluated all five models trained via 5-fold cross-validation on the Charité dataset and report the mean and standard deviation of the slide-AUROC scores on the LMU cohort.

\section{Additional results} \label{app:results}
This section provides additional experimental results, supplementing the results presented in the main paper.

\subsection{Performance on the external LMU cohort} \label{app:lmu_results}
Table \ref{tab:lmu-performance} shows the performance on the external hold-out LMU cohort. 

\begin{table*}[h!]
    \caption{Anomaly detection performance on the stomach and colon LMU cohort. We only report slide-AUROC scores as no pixel-wise annotations were available for this cohort.}
    \scriptsize
    \label{tab:lmu-performance}
    \centering
    \begin{tabular}{llcc}\toprule
     & &   \multicolumn{1}{c|}{\textbf{Stomach}} & \multicolumn{1}{c}{\textbf{Colon}} \\
    Model & Diagnosis Group & \multicolumn{1}{c}{slide-AUROC}  & \multicolumn{1}{|c}{slide-AUROC}  \\
 \midrule
Self-supervision w/ kNN &                              &          $88.6 \pm 0.1$  &        $84.44 \pm 0.61$  \\
                                     & \tiny{Neoplastic, malignant} & \tiny{$87.91 \pm 0.15$}  &  \tiny{$97.89 \pm 0.1$} \\
                                     &    \tiny{Neoplastic, other} & \tiny{$99.15 \pm 0.09$}  &  \tiny{$100.0 \pm 0.0$}  \\
                                     &          \tiny{Inflammation} & \tiny{$90.24 \pm 0.64$}  & \tiny{$78.72 \pm 0.98$} \\
                                     &  \tiny{Other}&    \tiny{-}  &  \tiny{$23.6 \pm 0.96$} \\
                                     &                \tiny{w/o pseudomelanosis and collagenous colitis} &  \tiny{-}  & \tiny{$90.05 \pm 0.54$}  \\
                                     \midrule
             Self-supervision w/ OCC &                              &        $89.92 \pm 0.85$ &                    $87.43 \pm 0.61$  \\
                                     & \tiny{Neoplastic, malignant} & \tiny{$89.28 \pm 0.98$} & \tiny{$97.79 \pm 0.25$} \\
                                     &    \tiny{Neoplastic, other} & \tiny{$96.67 \pm 0.65$}  &  \tiny{$100.0 \pm 0.0$}  \\
                                     &          \tiny{Inflammation} & \tiny{$89.11 \pm 0.85$}  & \tiny{$83.27 \pm 0.85$}  \\
                                     &  \tiny{Other}&    \tiny{-}  & \tiny{$34.92 \pm 1.35$}  \\
                                     &                \tiny{w/o pseudomelanosis and collagenous colitis} & \tiny{-}  &  \tiny{$93.3 \pm 0.61$}  \\
                                     \midrule
Outlier Exposure w/ BCE  &                              &         $94.5 \pm 0.93$ &                         $85.88 \pm 0.94$  \\
                                     & \tiny{Neoplastic, malignant} & \tiny{$94.77 \pm 0.88$}  & \tiny{$95.02 \pm 0.37$}  \\
                                     &    \tiny{Neoplastic, other} & \tiny{$99.47 \pm 0.27$}  & \tiny{$98.57 \pm 0.24$}  \\
                                     &          \tiny{Inflammation} & \tiny{$90.08 \pm 2.24$}  & \tiny{$82.53 \pm 1.42$}  \\
                                     &  \tiny{Other}&    \tiny{-} & \tiny{$30.48 \pm 2.86$}  \\
                                     &                \tiny{w/o pseudomelanosis and collagenous colitis} & \tiny{-}  & \tiny{$91.71 \pm 0.57$} \\
\bottomrule
\end{tabular}
\end{table*}

\subsection{Detection performance for each distinct diagnosis in the Charité cohort} \label{app:per_diagnosis_results}
We report patch-AUROC and slide-AUROC for each distinct diagnosis in our Charité cohort in Table \ref{tab:diagnosis-results-stomach} (stomach) and Table \ref{tab:diagnosis-results-colon} (colon). 
\begin{table}[h]
    \centering
    \scriptsize
    \caption{Patch-AUROC and slide-AUROC scores for each distinct diagnosis in the stomach Charité cohort for the OE model. The scores are computed by assessing only the anomalous data from the respective diagnosis type and all frequent findings.}
    \label{tab:diagnosis-results-stomach}
    \begin{tabular}{llcc}\toprule
Diagnosis Group & Diagnosis  &  patch-AUROC &  slide-AUROC \\
\midrule
Neoplastic, malignant & Gastric adenocarcinoma, NFS & $95.54 \pm 0.39$ & $98.14 \pm 0.53$ \\
 & Gastric adenocarcinoma, signet-ring-cell & $97.66 \pm 0.20$ & $99.18 \pm 0.22$ \\
 & Gastric adenocarcinoma, hepatoid & $96.63 \pm 0.48$ & $99.27 \pm 0.47$ \\
 & Adenosquamous carcinoma & $97.52 \pm 0.18$ & $99.79 \pm 0.29$ \\
 & Squamous cell carcinoma & $97.67 \pm 0.18$ & $99.76 \pm 0.27$ \\
 & Undifferentiated carcinoma & $99.63 \pm 0.16$ & $99.79 \pm 0.29$ \\
 & Isolated lymphangiosis carcinomatosa & $94.02 \pm 0.53$ & $99.27 \pm 0.61$ \\
 & Neuroendocrine tumor & $91.21 \pm 0.86$ & $93.70 \pm 0.90$ \\
 & MINEN & $98.29 \pm 0.22$ & $99.16 \pm 0.40$ \\
 & Neuroendocrine carcinoma & $98.40 \pm 0.23$ & $96.39 \pm 0.60$ \\
 & B-cell lymphoma, NFS & $93.26 \pm 0.43$ & $97.61 \pm 0.59$ \\
 & Marginal zone lymphoma & $91.12 \pm 0.50$ & $95.52 \pm 0.77$ \\
 & Gastrointestinal stromal tumor & $98.10 \pm 0.25$ & $99.63 \pm 0.30$ \\
 & Undifferentiated sarcoma & $97.80 \pm 0.23$ & $99.79 \pm 0.29$ \\
 & Metastasis, Adenocarcinoma of pancreas & $96.40 \pm 0.39$ & $98.23 \pm 0.81$ \\
 & Metastasis, Adenocarcinoma of breast & $91.58 \pm 0.60$ & $97.79 \pm 0.88$ \\
 & Metastasis, Adenocarcinoma of lung & $96.63 \pm 0.22$ & $99.79 \pm 0.29$ \\
 & Metastasis, Urothelial carcinoma & $99.05 \pm 0.18$ & $99.79 \pm 0.29$ \\
 & Metastasis, Melanoma & $92.17 \pm 0.56$ & $95.00 \pm 0.72$ \\
\midrule
Neoplastic, other & Tubular adenoma & $94.47 \pm 0.25$ & $99.51 \pm 0.38$ \\
 & Tubulovillous adenoma & $98.25 \pm 0.30$ & $99.69 \pm 0.29$ \\
 & Adenoma, oxyntic type & $73.33 \pm 2.13$ & $64.42 \pm 3.70$ \\
 & Adenoma, foveolar type & $86.69 \pm 0.84$ & $73.03 \pm 1.82$ \\
 & Fundic gland polyp & $88.43 \pm 1.24$ & $98.95 \pm 0.76$ \\
 & Hyperplastic polyp & $92.87 \pm 0.51$ & $97.04 \pm 0.64$ \\
 & Leiomyoma & $95.18 \pm 0.54$ & $99.16 \pm 0.61$ \\
 & Peutz-Jeghers polyp & $73.00 \pm 1.06$ & $87.22 \pm 1.52$ \\
\midrule
Inflammation & Ulcer & $96.73 \pm 0.26$ & $95.26 \pm 0.89$ \\
 & Lymphocytic gastritis & $69.23 \pm 3.00$ & $72.93 \pm 2.70$ \\
\midrule
Other & Helminthosis & $99.85 \pm 0.11$ & $99.79 \pm 0.29$ \\
 & Xanthoma & $86.80 \pm 1.09$ & $92.24 \pm 0.92$ \\
 & Calcinosis & $92.25 \pm 0.84$ & $97.71 \pm 0.31$ \\
 & Pancreatic heterotopia & $96.35 \pm 0.49$ & $95.84 \pm 0.71$ \\
\bottomrule
\end{tabular}
\end{table}

\begin{table}[h!]
    \centering
    \scriptsize
    \caption{Patch-AUROC and slide-AUROC scores for each distinct diagnosis in the colon Charité cohort for the OE model. The scores are computed by assessing only the anomalous data from the respective diagnosis type and all frequent findings.}
    \label{tab:diagnosis-results-colon}
    \begin{tabular}{llll}\toprule
Diagnosis Group & Diagnosis  &  patch-AUROC &  slide-AUROC \\
\midrule
Neoplastic, malignant & Colorectal adenocarcinoma & $96.14 \pm 0.27$ & $98.53 \pm 0.67$ \\
 & Squamous cell carcinoma & $95.63 \pm 0.34$ & $97.49 \pm 1.51$ \\
 & Neuroendocrine tumor & $95.03 \pm 0.37$ & $85.68 \pm 1.06$ \\
 & Undifferentiated sarcoma & $98.85 \pm 0.26$ & $100.0 \pm 0.00$ \\
 & Metastasis, Melanoma & $98.03 \pm 0.41$ & $99.85 \pm 0.23$ \\
\midrule
Neoplastic, other & Tubular adenoma, high grade & $89.56 \pm 0.46$ & $90.23 \pm 1.95$ \\
 & Tubulovillous adenoma, high grade & $95.87 \pm 0.46$ & $99.35 \pm 0.40$ \\
 & Villous adenoma, high grade & $88.27 \pm 1.16$ & $93.39 \pm 1.93$ \\
 & Leiomyoma & $97.83 \pm 0.24$ & $100.0 \pm 0.00$ \\
\midrule
Inflammation & Crohn's disease & $91.86 \pm 0.46$ & $98.44 \pm 0.65$ \\
 & Ulcerative colitis & $87.95 \pm 0.66$ & $95.20 \pm 1.45$ \\
 & Acute and chronic colitis & $91.28 \pm 0.47$ & $94.19 \pm 1.75$ \\
 & Ischemic colitis & $96.03 \pm 0.41$ & $99.49 \pm 0.64$ \\
 & Collagenous colitis & $70.19 \pm 0.68$ & $83.82 \pm 3.07$ \\
 & Lymphocytic colitis & $51.71 \pm 0.38$ & $79.39 \pm 0.71$ \\
 & Ulcer & $96.84 \pm 0.31$ & $99.19 \pm 0.46$ \\
 & Inflammatory pseudopolyp & $92.78 \pm 0.44$ & $98.75 \pm 0.62$ \\
 & Pseudomembranous colitis & $95.19 \pm 0.64$ & $97.21 \pm 1.00$ \\
 & Sevelamer-induced colitis & $73.79 \pm 1.04$ & $85.99 \pm 5.34$ \\
\midrule
Other & Helminthosis & $78.50 \pm 1.49$ & $95.90 \pm 1.24$ \\
 & Intestinal spirochaetosis & $40.87 \pm 1.43$ & $23.56 \pm 2.58$ \\
 & Pseudomelanosis Coli & $28.86 \pm 0.90$ & $20.28 \pm 3.36$ \\
\bottomrule
\end{tabular}
\end{table}


\subsection{Artifacts vs.\ anomalies} \label{app:artifacts}
We also annotated tissue- and processing artifacts on anomaly slides, checking whether they might cause ``Clever Hans'' effects \cite{lapuschkin2019unmasking,kauffmann2020clever} where a slide receives a high anomaly score for the wrong reasons, i.e.\ for potential artifacts possibly driving high anomaly scores of infrequent cases. However, compared to annotated anomaly regions, the artifact regions received significantly lower anomaly scores. The artifact regions had patch-AUROCs of 80.89\% (colon) and 75.64\% (stomach) while anomaly regions received patch-AUROCs of 90.47\% (colon) and 91.37\% (stomach). This shows that artifacts did not have a major influence on our results.

\subsection{Anomaly heatmaps} \label{app:add_heatmaps}
\begin{figure}[h!]
    \centering
    \includegraphics[width=0.75\textwidth]{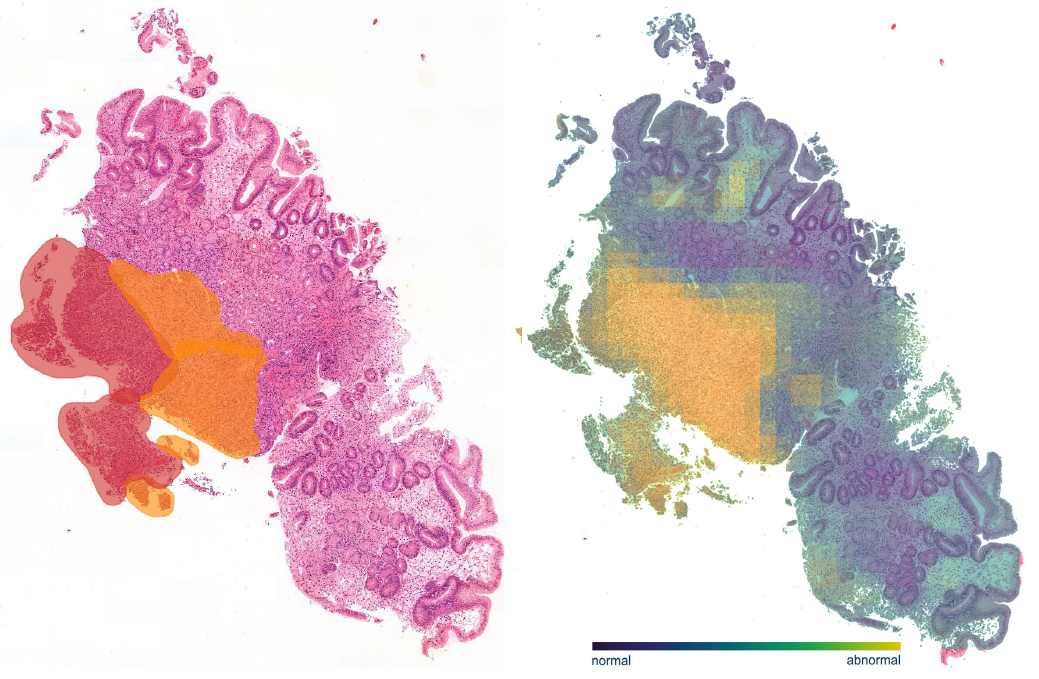}
    \caption{Stomach tissue with metastatic infiltrates of a melanoma. A complete tissue cut with pathologists' annotations is shown on the left. The corresponding anomaly heatmap is shown on the right.}
    \label{fig:stomach-melanoma}
\end{figure}

\begin{figure}[H]
    \centering
    \includegraphics[width=.85\textwidth]{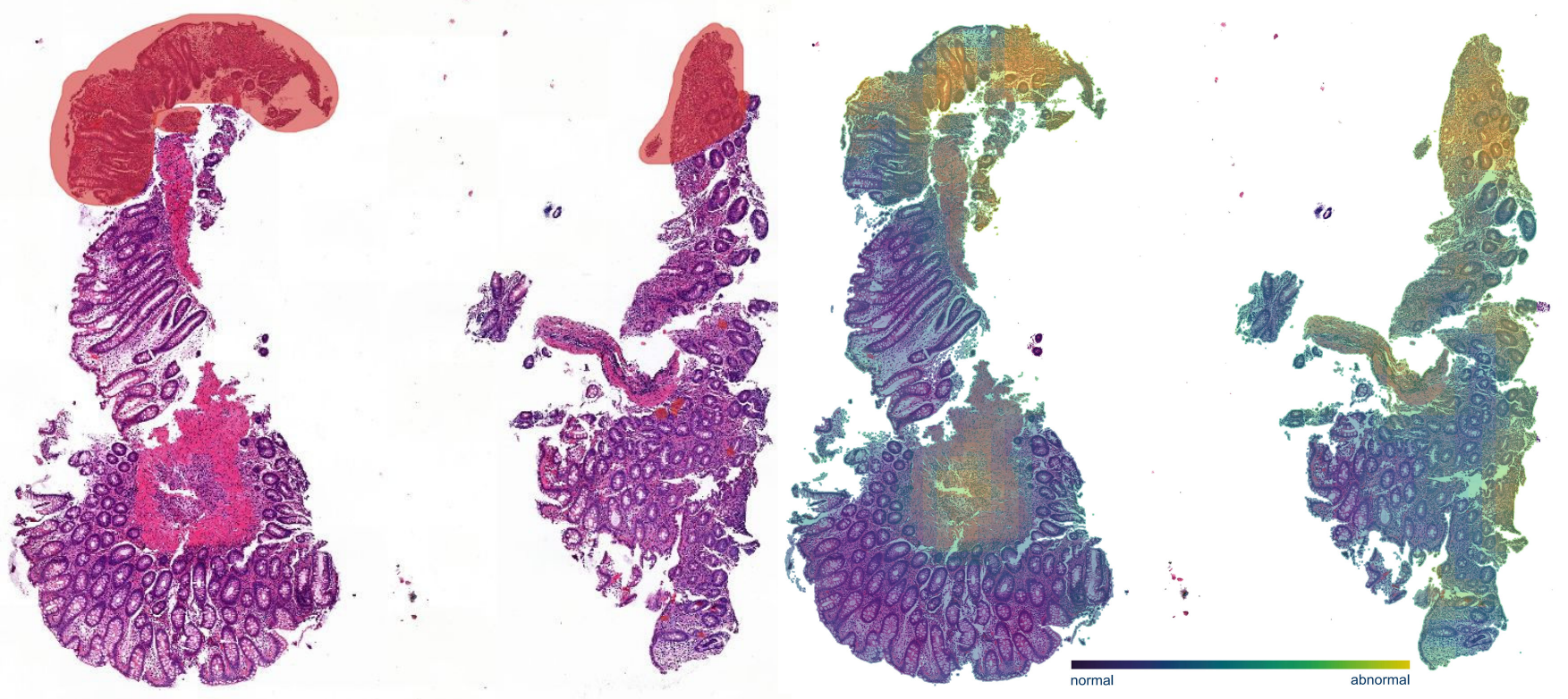}
    \caption{Colon tissue with focal unspecific inflammatory changes. A complete tissue cut with pathologists' annotations is shown on the left. The corresponding anomaly heatmap is shown on the right. As transitions of normal tissue areas to abnormal tissue areas are vague, annotations represent the most anomalous areas.}
    \label{fig:colon-inflammation}
\end{figure}

\begin{figure}[h!]
    \centering
    \includegraphics[width=\textwidth]{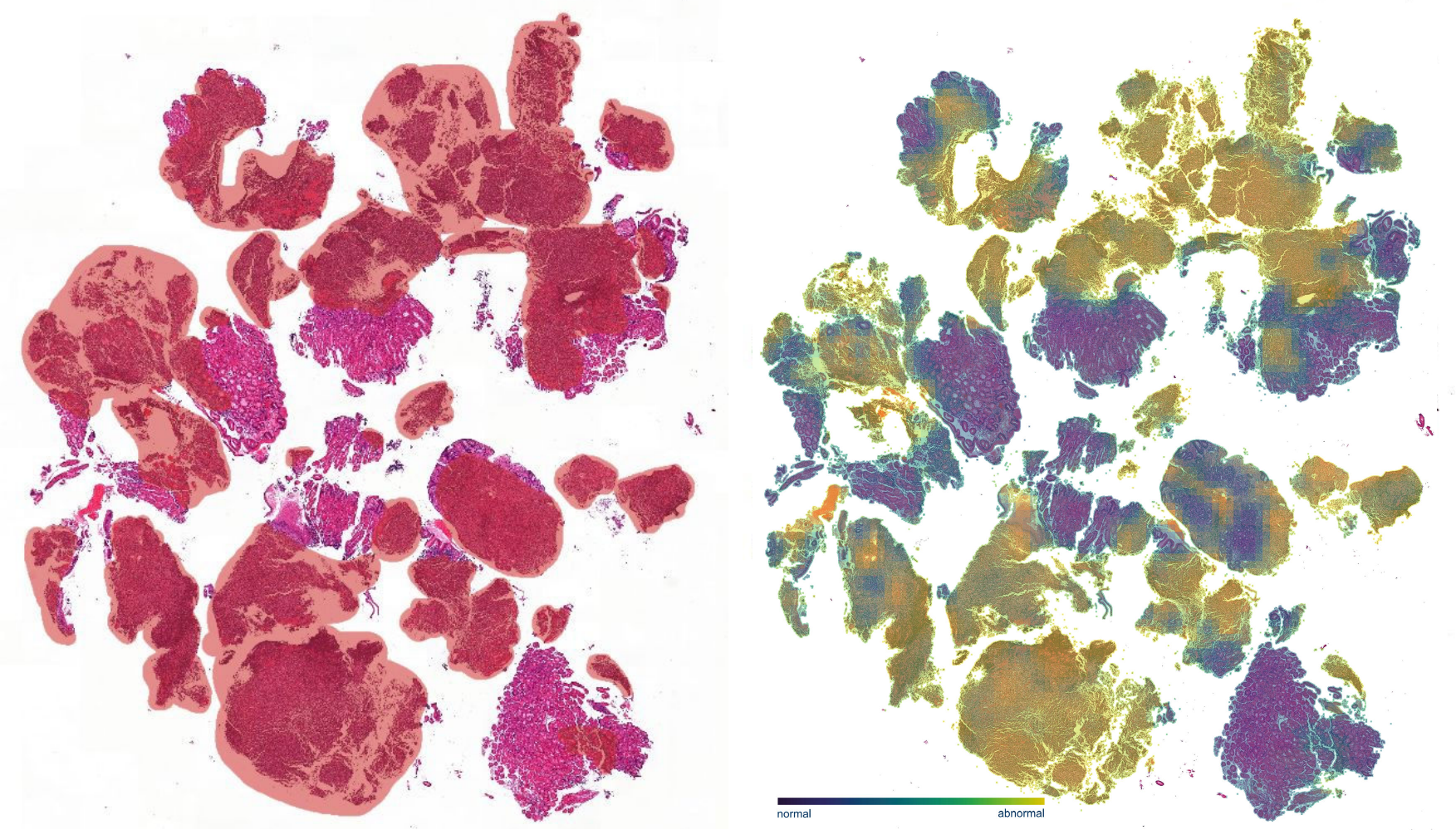}
    \caption{Stomach tissue with infiltrates of marginal zone lymphoma. A complete tissue cut with pathologists' annotations is shown on the left. The corresponding anomaly heatmap is shown on the right.}
    \label{fig:stomach-lymphoma}
\end{figure}

\begin{figure}[H]
    \centering
    \includegraphics[width=\textwidth]{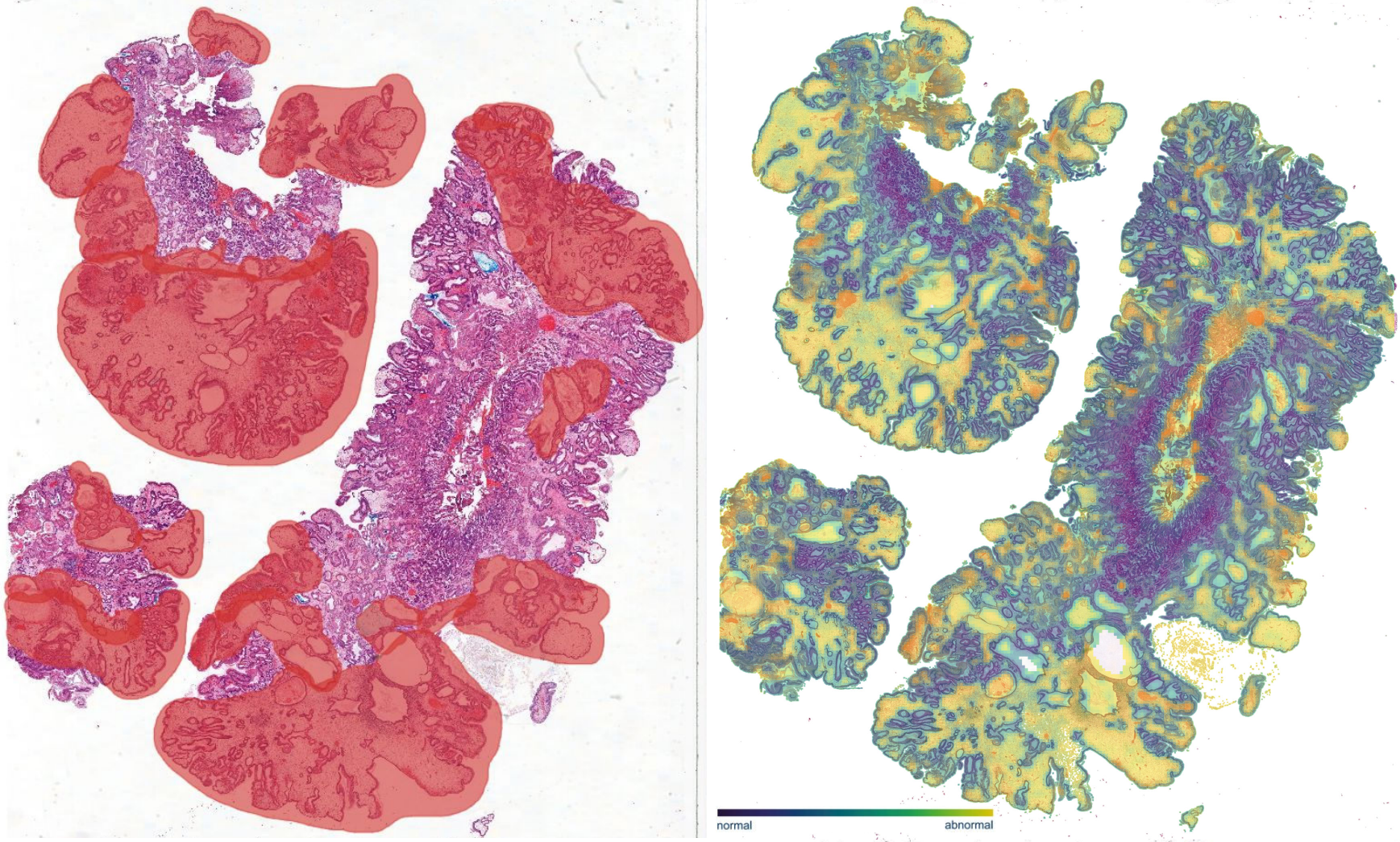}
    \caption{Hyperplastic polyp of the stomach. A complete tissue cut with pathologists' annotations is shown on the left. The corresponding anomaly heatmap is shown on the right. As transitions of normal tissue areas to abnormal tissue areas are vague, annotations represent the most anomalous areas.}
    \label{fig:stomach-stomach-hyperplasticpolyp}
\end{figure}

\begin{figure}[H]
    \centering
    \includegraphics[width=\textwidth]{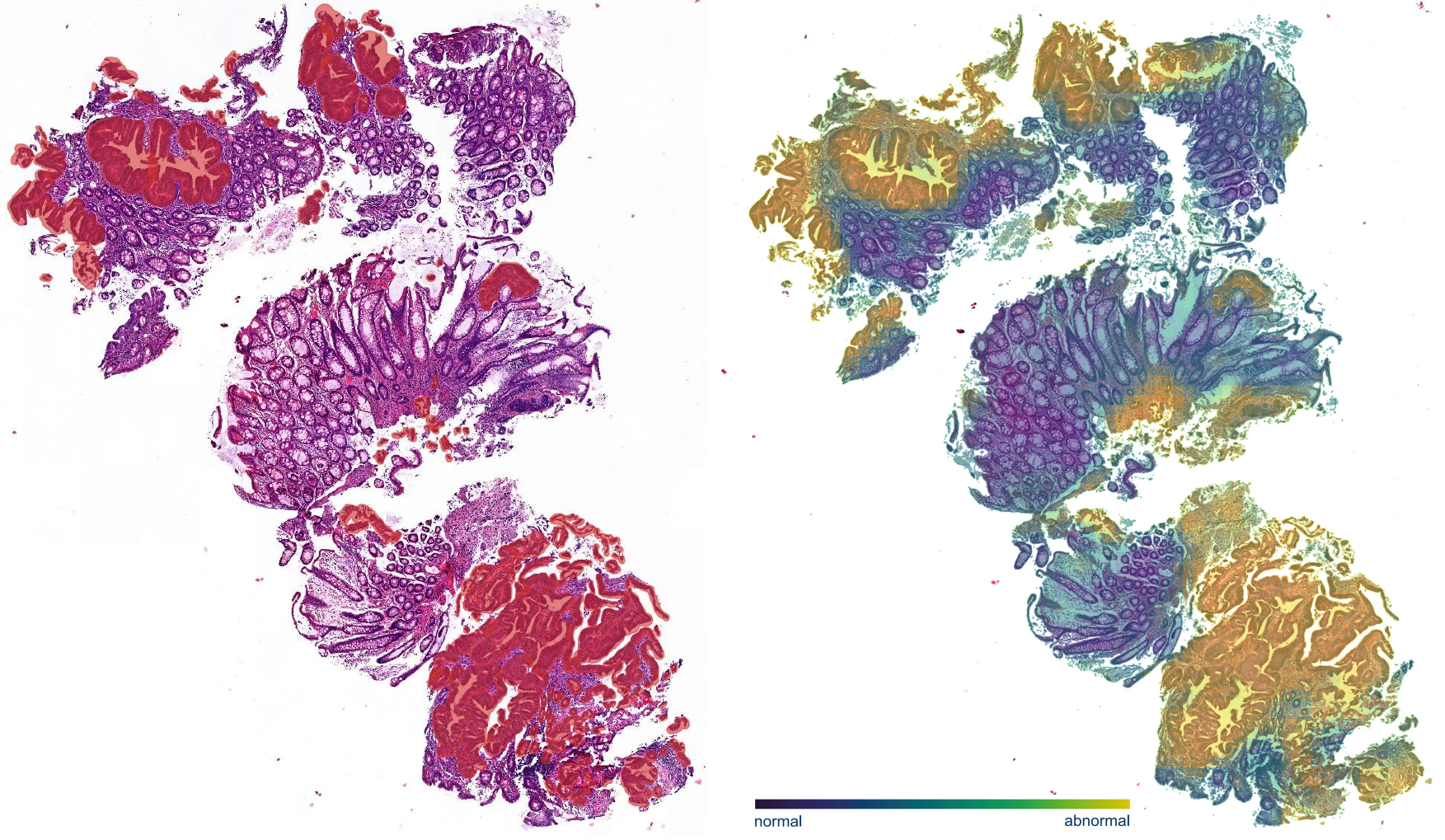}
    \caption{Colon tissue with infiltrates of a colorectal adenocarcinoma. A complete tissue cut with pathologists' annotations is shown on the left. The corresponding anomaly heatmap is shown on the right.}
    \label{fig:colon-adenocarcinoma}
\end{figure}

\begin{figure}[H]
    \centering
    \includegraphics[width=\textwidth]{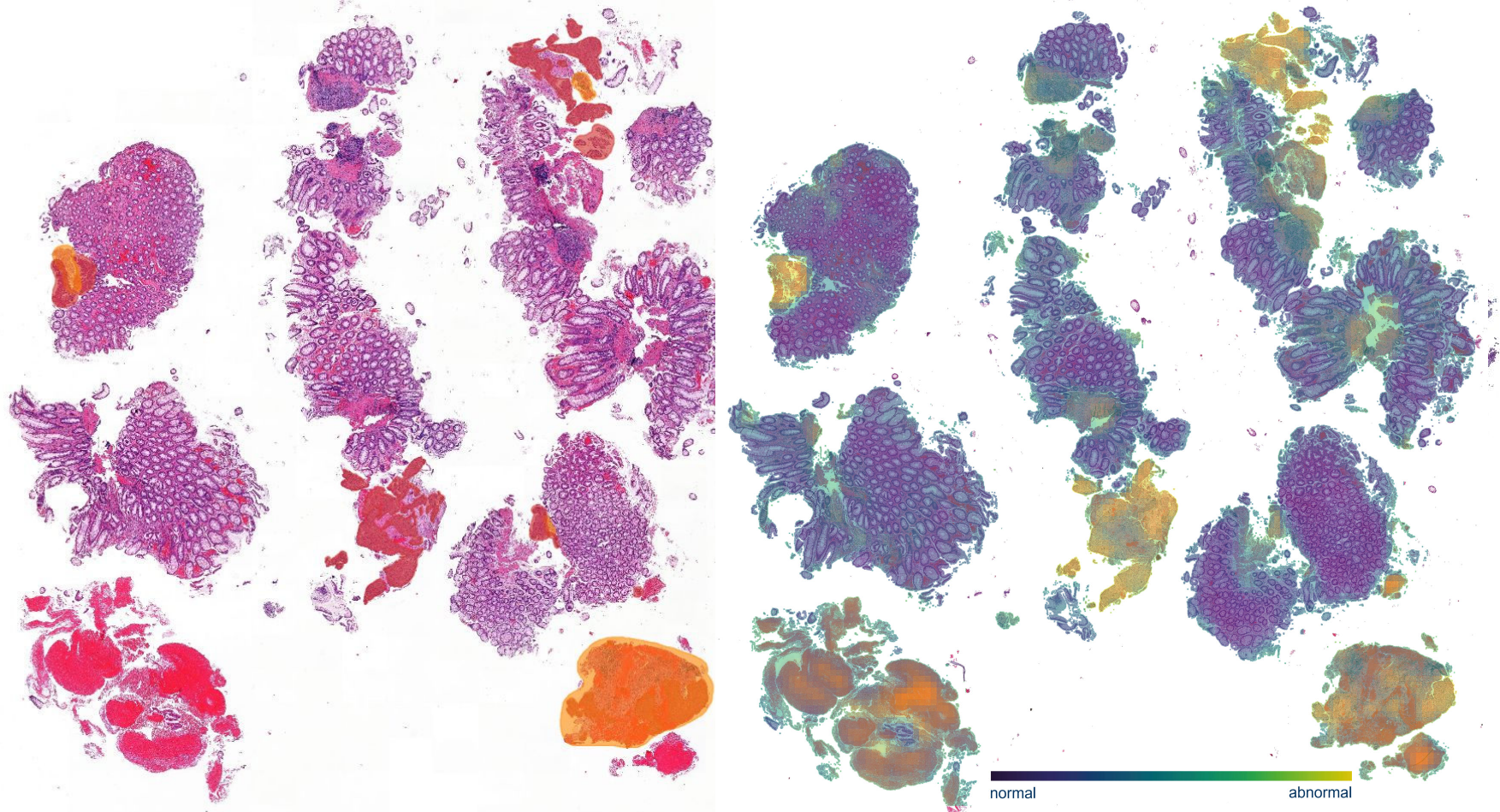}
    \caption{Colon tissue with infiltrates of a squamous cell carcinoma. A complete tissue cut with pathologists' annotations is shown on the left. The corresponding anomaly heatmap is shown on the right.}
    \label{fig:colon-squamous}
\end{figure}

We also investigated the model performance qualitatively by presenting heatmaps of our AD model to pathologists. In the main paper, we provide selected excerpts of heatmaps to visually demonstrate the performance of our AD model. Here, in Figures \ref{fig:stomach-melanoma}--\ref{fig:colon-squamous}, we present additional heatmaps (right side) along with pixel-wise annotations by pathologists (left side). Areas annotated with red color indicate anomalous regions that define the final diagnosis. Areas annotated in yellow indicate other anomalous regions that do not directly define the final diagnosis (e.g. inflammatory changes adjacent to tumor tissue). We show complete tissue cuts of whole slide images.

\subsection{kNN with self-supervised models} \label{app:knn_results}
Any model that provides informative image representations can be used for AD by simply applying the k-nearest-neighbor (kNN) algorithm in representation space. This often results in state-of-the-art or competitive performance \cite{dn2, reiss2021panda, tack2020, muttenthaler2023improving}.

\begin{table}[h!]
    \centering
    \footnotesize
    \caption{Overview of self-supervised models that have been pretrained on a large collection of histopathology images. All methods maximize the similarity of two augmented views from the same image.}
    \label{tab:self-sup-models}
    \begin{tabular}{c|ccc}
         Name & Architecture & Method & Dataset  \\
         \midrule
         HIPT \cite{chen2022scaling} & VIT-S/16 & DINO & 33 cancer types, 104M 256×256 images \\
         R50 SimCLR BRCA \cite{chen2022self} & ResNet-50 & SimCLR &  TCGA-BRCA. \\
         Ciga et al. \cite{ciga2022self} & ResNet-18 & SimCLR & 23 non WSI and 35 WSI datasets \\
         RetCCL \cite{retccl} & ResNet-50 & contrastive learning & 32,000 WSIs \\
         CTranspath \cite{ctranspath} & SwinTransformer & mod. SimCLR & TCGA + PAIP
    \end{tabular}
\end{table}

Suitable models for our scenario, i.e., those pretrained with diverse self-supervised techniques on a large collection of histopathology images, are readily available online. Additionally, we trained our own self-supervised model with the SimCLR framework on the normal data.
Further, we fine-tuned the self-supervised models with a one-class loss \cite{reiss2021panda}. 

\begin{table*}[h!]
    \small
    \centering
    \caption{Anomaly detection performance with self-supervised models on the stomach and colon Charité cohort. All models use kNN for AD.}
    \label{tab:ssl-ablation}
    \begin{tabular}{cc|cc|cc}
    & & \multicolumn{2}{c}{Colon} &  \multicolumn{2}{c}{Stomach} \\
    Cohort & Method   & patch-AUROC & slide-AUROC & patch-AUROC & slide-AUROC \\
    \midrule
    Charité &                         R50 ImageNet &           54.65 &           66.42 &             61.11 &             73.94 \\
        & R50 SimCLR BRCA \cite{chen2022self} &           67.02 &           84.57 &             70.84 &             86.63 \\
        &      Ciga et al. \cite{ciga2022self} &           68.07 &           75.91 &             76.22 &             90.55 \\
        & Self-trained R18 SimCLR &           60.13 &           80.79 &             73.66 &             91.54 \\
        &         HIPT \cite{chen2022scaling} &           75.55 &           85.84 &             75.68 &             87.31 \\
        &                               RetCCL \cite{retccl} &           77.95 &           86.86 &             84.65 &             93.64 \\
        &                           CTransPath \cite{ctranspath} &            \textbf{84.30} &           \textbf{90.39} &             \textbf{87.16} &             \textbf{95.09} \\
    \end{tabular}
\end{table*}

We investigated the AD performance of those models with kNN. We only report the best model in the main paper, while, in this section, we show the results for all other self-supervised models. Table \ref{tab:self-sup-models} displays an overview of the considered models, and Table \ref{tab:ssl-ablation} shows the results for the Charité cohort. We observed that the CTranspath model  \cite{ctranspath} outperforms all other models on both tissue types. Table \ref{tab:panda-ablation} shows the models after finetuning with a one-class loss. Finetuning improves the patch-AUROC for some models but does not bring significant improvements on the slide level.

\begin{table*}[h!]
    \small
    \centering
    \caption{Anomaly detection performance with self-supervised models on the stomach and colon Charité cohort after finetuning with a one-class loss. All models use kNN for AD.}
    \label{tab:panda-ablation}
    \begin{tabular}{cc|cc|cc}
    & & \multicolumn{2}{c}{Colon} &  \multicolumn{2}{c}{Stomach} \\
    Cohort & Method   & patch-AUROC & slide-AUROC & patch-AUROC & slide-AUROC \\
    \midrule
    Charité &                         R50 ImageNet &           54.03 &            63.30 &             61.75 &             68.74 \\
        & R50 SimCLR BRCA \cite{chen2022self} &           66.05 &           82.83 &             70.27 &             84.08 \\
        &      R18 SimCLR \cite{ciga2022self} &           70.69 &           72.88 &             77.83 &              83.80 \\
        & HIPT \cite{chen2022scaling} &           76.98 &           84.82 &             80.49 &             87.41 \\
        &                               RetCCL \cite{retccl} &           82.72 &           85.89 &             86.51 &             87.18 \\
        &                           CTransPath \cite{ctranspath} &           \textbf{86.24} &           \textbf{88.93} &             \textbf{89.56} &             \textbf{93.52} \\
    \end{tabular}
\end{table*}

\newpage

\subsection{Autoencoder}  \label{app:autoencoder_results}
Many of the discussed previous works for AD in histopathology used reconstruction-based methods to detect anomalies \cite{pocevivciute2021unsupervised, litjens20181399, stepec2021unsupervised, zehnder2022multiscale}. Therefore, we also investigated the performance of an autoencoder model on our dataset. As mentioned in the related work, an autoencoder is trained on the normal data to compress the image into a low-dimensional representation and then reconstructs the image from that latent representation. The main idea is that the model has a larger reconstruction error on anomalies, as it had only been trained on normal data, thus making the reconstruction error an anomaly score. 

We trained a simple autoencoder with a bottleneck dimension of 512 and 6 blocks with 2 convolutional layers each in the encoder and decoder, respectively. Table \ref{tab:ae-ablation} shows the performance of this autoencoder model.

\begin{table*}[h!]
    \small
    \caption{Anomaly detection performance of a simple autoencoder model compared to our OE based model.}
    \label{tab:ae-ablation}
    \centering
    \begin{tabular}{ll|cc|cc}
    & &   \multicolumn{2}{|c|}{\textbf{Colon}} &  \multicolumn{2}{c}{\textbf{Stomach}} \\
     Cohort & Method & \multicolumn{1}{|c}{patch-AUROC} & \multicolumn{1}{c|}{slide-AUROC} & patch-AUROC & slide-AUROC \\
    \midrule
   Charité &      Autoencoder &           59.67 &           65.58 &             51.33 &             63.73 \\
        & Outlier Exposure &           \textbf{90.38} &           \textbf{91.49} &              \textbf{90.90} &              \textbf{94.40} \\
    LMU &      Autoencoder &               - &           55.54 &                 - &             55.12 \\
        & Outlier Exposure &               - &           \textbf{84.58} &                 - &             \textbf{93.07} \\
\bottomrule
\end{tabular}
\end{table*}

We can observe that the performance of the autoencoder model is poor compared to our OE approach. Especially for the stomach cohort, the performance is not considerably better than random (51.33 patch-AUROC). We could potentially improve the autoencoder model with an additional adversarial or perceptual loss. However, these improvements usually only result in small performance improvements \cite{bergmann2019mvtec}. In contrast to industry defect datasets \cite{bergmann2019mvtec}, where there is typically only small variation within the normal data, the normal data in our case has a great intrinsic variation.

\subsection{Multi-scale considerations} \label{app:multiscale}
Even for pathologists, some pathological patterns are barely detectable on individual patches of some fixed size or may look very similar to frequently found and healthy tissue characteristics. We experienced this issue in some anomalous findings.

\begin{figure}[H]
    \centering
    \includegraphics[width=\textwidth]{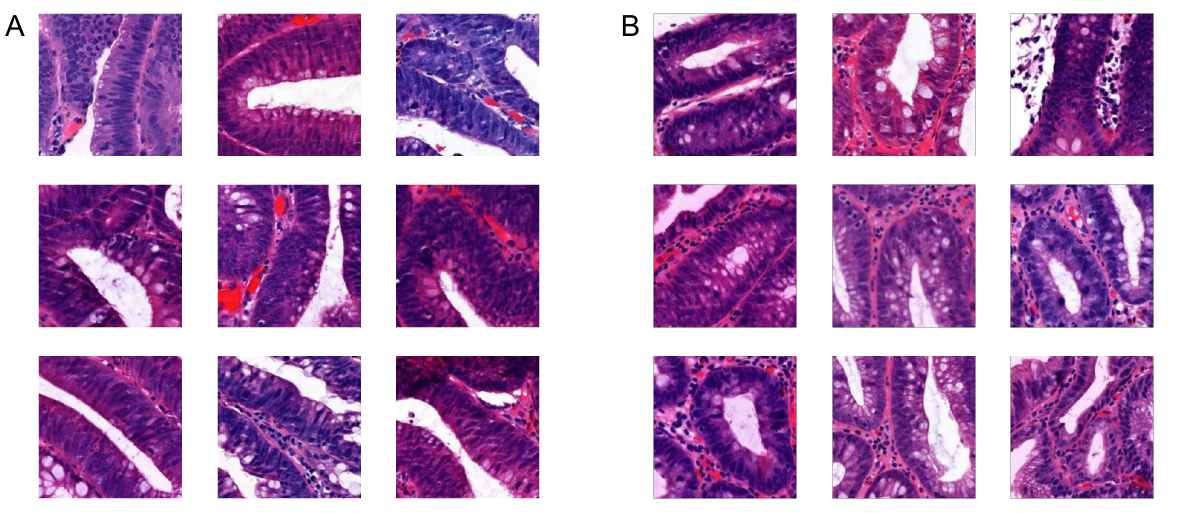}
    \caption{Similarity of patches from low-grade adenomas (left) and patches of crypts of regular colon mucosa (right). We generated embeddings from patches of annotated low-grade adenoma regions with the CTransPath model and then retrieved similar patches in normal tissue regions through a nearest neighbor search.}
    \label{fig:normalvsadenoma}
\end{figure}

For example, histomorphological changes from normal colon tissue to adenomas with low-grade or high-grade cellular changes and even to adenocarcinoma occur on a continuous scale. We noticed that findings from different stages along that continuum can look very similar on a patch level, as exemplified in Figure \ref{fig:normalvsadenoma}. To set a clear distinction between non-neoplastic colon tissue and high-grade epithelial changes/adenocarcinoma, which are crucial to detect, we decided to exclude low-grade adenomas from our training data. This enabled us to get high detection rates for potentially cancerous findings, while at the same time preserving low anomaly scores for low-grade adenomas because of their high similarity to deep crypt characteristics in regular colon mucosa on a patch level.

Similarly, tissue changes consistent with leiomyoma can often not be differentiated from normal smooth muscle tissue on a patch level. Here, the size and context of anomalous regions are needed to make an accurate diagnosis in clinical practice. However, as smooth muscle tissue of deep biopsies in our dataset tends to get slightly higher anomaly scores than mucosal tissue, the leiomyomas in our dataset received notable slide anomaly scores after aggregation. 

As shown, tissue context beyond patch size is often helpful in detecting anomalies. Architectural changes in colon mucosa during inflammation or post-inflammation are another example of this. Multi-scale approaches, integrating tissue contexts at different magnification levels, might be a promising avenue to improve the detection performance for these kinds of anomalies, which we plan to explore in future work.

\section{Ablation experiments}  \label{app:ablations}
We performed ablation experiments to investigate the effects of the different building blocks in our OE model.

\subsection{Varying normal data size}  \label{app:dataset_size_ablation}
We used a large number of frequent finding slides ($\approx$ 1000 each for stomach and colon) for training our AD model. In this ablation, we investigated how much training data is necessary for the AD model to generalize to unseen anomalies and whether collecting more training slides would result in a significant performance gain. To test this scaling behavior, we varied the number of slides that we used during training and observed the resulting AD performance on the test set. 

\begin{figure}[h!]
    \centering
    \small
    \begin{subfigure}[b]{0.49\textwidth}
         \centering
         \includegraphics[width=\textwidth]{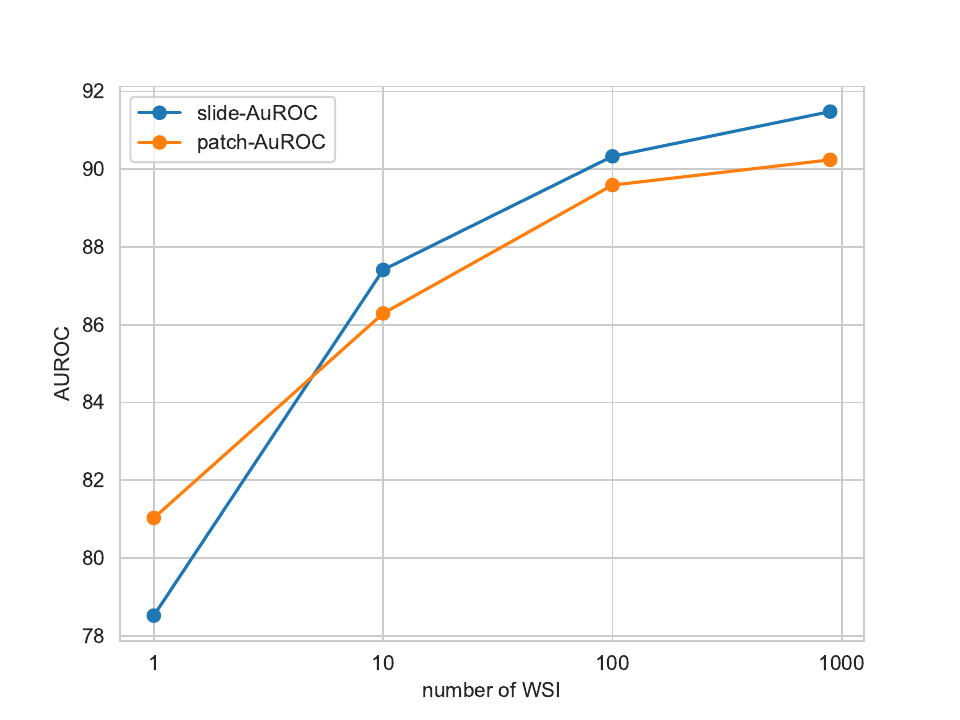}
         \subcaption{Colon}
     \end{subfigure}
     \begin{subfigure}[b]{0.49\textwidth}
         \centering
         \includegraphics[width=\textwidth]{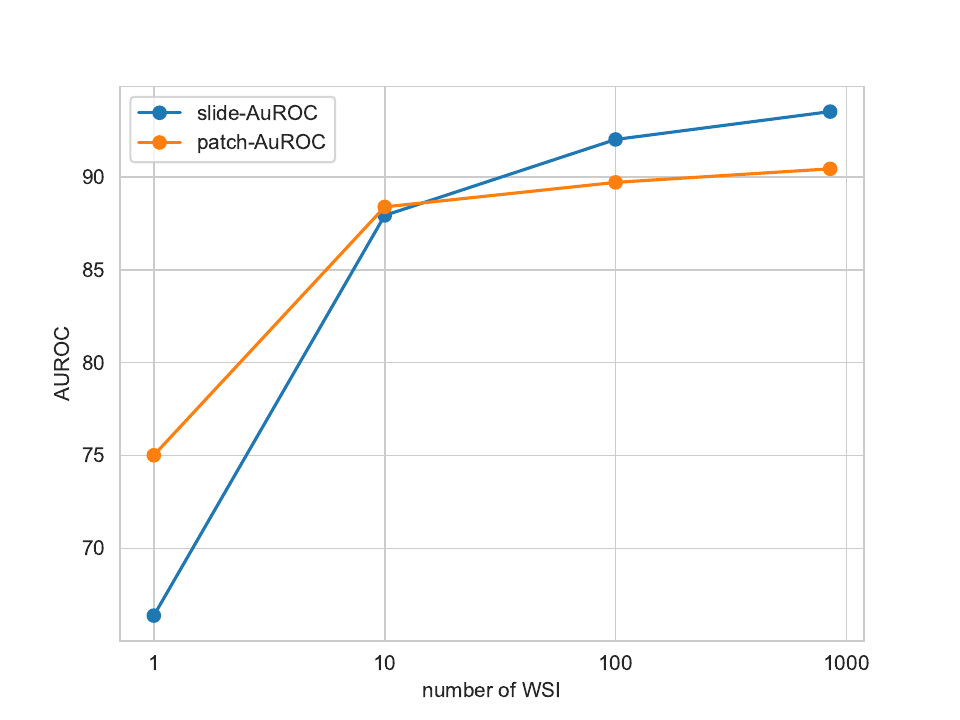}
         \subcaption{Stomach}
     \end{subfigure}
    \caption{Slide-AUROC and patch-AUROC scores with respect to the number of WSIs of frequent findings in the training set. The models were all trained for the same amount of steps.}
    \label{fig:normal-data-size}
\end{figure}

We report results for 1, 10, and 100 randomly sampled slides, as well as the full training dataset. We trained all models for the same amount of iterations. Figure \ref{fig:normal-data-size} shows the performance on the Charité cohort with the x-axis being on a logarithmic scale.  

The plots show patch-AUROC and slide-AUROC scores with respect to the number of WSIs used during training. As expected, the results highlight that more WSIs in the training data lead to an improved AD performance. Having more data of frequent findings available at training time, allows the model to see more patterns and therefore enables better generalization to unseen data. We observed that the performance started to saturate at around 100 slides. Adding a magnitude of more training slides only marginally improved the performance. This indicates that 100 slides are already sufficient to capture most of the patterns in the training data. Collecting this amount of slides from routine diagnostics is practical and the scaling behavior is consistent between colon and stomach. Therefore, we believe that AD can be adapted well for other tissues with limited slide collection needs.

\subsection{OE datasets}  \label{app:oe_ablation}

In the following, we show that the selection of suitable OE data is nontrivial and of utter importance for strong generalization to true anomalies. Table \ref{tab:oe-dataset-ablation} shows the performance for different OE datasets, which we will elaborate on in the following. Throughout all ablations, we used the same train/test split and do not report results on full 5-fold cross-validation. As the standard deviation on the 5-fold cross-validation is low, significant performance differences are also evident with this reduced evaluation scheme.

\begin{table*}[h!]
    \small
    \caption{AD performance for different OE datasets on the Charité cohort. Further information about the datasets is given in the following paragraphs.}
    \label{tab:oe-dataset-ablation}
    \centering
    \begin{tabular}{ll|cc|cc}
    & &   \multicolumn{2}{|c|}{\textbf{Colon}} &  \multicolumn{2}{c}{\textbf{Stomach}} \\
      & Dataset & \multicolumn{1}{|c}{patch-AUROC} & \multicolumn{1}{c|}{slide-AUROC} & patch-AUROC & slide-AUROC \\
    \midrule
        & ImageNet-1K & 50.72 & 52.17 & 55.91 & 75.21 \\
        & TCGA & 78.84 & 82.73 & 80.73 & 88.80 \\
        &  Full OE Set & \textbf{90.38} & \textbf{91.49} & \textbf{90.90} & \textbf{94.40} \\
        & \hspace{2mm} Charité OE mixed & 86.09 & 88.48 & 88.34 & 93.53 \\
        & \hspace{4mm} Kidney subset & 85.13 & 86.78 & 85.47 & 88.32 \\
        & \hspace{4mm} Breast subset & 84.98 & 88.96 & 87.13 & 89.90 \\
        & \hspace{2mm} Small Intestine & \textbf{90.41} & \textbf{91.74} & 89.54 & 92.44 \\
        & \hspace{2mm} Stomach & \textbf{90.73} & \textbf{91.61} & - & - \\
        & \hspace{2mm} Colon & - & - & 88.48 & 90.99 \\
\bottomrule
\end{tabular}
\end{table*}

\subsubsection*{OE using natural images}

On natural image AD benchmarks (e.g., one vs.~rest with CIFAR10), OE with a diverse set of natural images is most effective \cite{hendrycks2019deep, ruff2020rethinking, liznerski2022exposing}.
In a first experiment, we investigated if this also holds in the histopathology regime, i.e., if a diverse OE set of natural images already suffices for strong generalization to true anomalies. The data distribution of natural images strongly differs from our medical images; therefore, we expected generalization to true anomalies to be challenging. We used the same training setup as for our main results but exchanged the OE dataset with the popular ImageNet-1K dataset \cite{imagenet} containing a set of 1000 image classes.

After training our model, we observed that the binary cross-entropy loss rapidly decreased after the first optimization steps. This shows that the model can easily differentiate between histopathology and natural images. After training, we evaluated our model on the test set, which yielded a poor performance of 50.72 patch-AUROC (colon) and 55.91 patch-AUROC (stomach) (see Table \ref{tab:oe-dataset-ablation}). The binary classification between natural images and our histopathology images is trivial in this case, and therefore the model does not learn features that generalize to real anomalies.

\subsubsection*{OE using TCGA data}

\begin{table}[h!]
    \caption{Statistics of TCGA OE data set. Patches were created at 20x magnification with patch size 340x340 pixels.}
    \label{tab:tcga-outlier-exposure-dataset}
    \centering
    \small
    \begin{tabular}{llrr}
        Study & Description & Slides & Patches \\
        \midrule
        BRCA & Breast invasive carcinoma            &  50 &  265,017 \\
        CHOL & Cholangiocarcinoma                   &  39 &  433,769 \\
        ESCA & Esophageal carcinoma                 &  50 &  338,414 \\
        LUAD & Lung adenocarcinoma                  &  49 &  336,924 \\
        PAAD & Pancreatic adenocarcinoma            &  48 &  332,582 \\
        PRAD & Prostate adenocarcinoma              &  50 &  372,134 \\
        UCEC & Uterine Corpus Endometrial Carcinoma &  50 &  490,544 \\
        \midrule
        Total &                                     & 336 & 2,569,384 \\
        \bottomrule
    \end{tabular}
\end{table}

Previous studies suggest that OE becomes more effective the more similar the OE data are to the normal data \cite{goyal2020drocc}, as this forces the model to learn a tighter decision boundary around the normal data. To obtain an OE dataset closer to our normal colon and stomach slides, we collected publicly available histopathology images from 7 selected studies of The Cancer Genome Atlas (TCGA\footnote{\url{https://www.cancer.gov/tcga}}): BRCA, CHOL, ESCA, LUAD, PAAD, PRAD, UCEC. The studies were chosen for some (however distant) morphological proximity to colon and stomach tissue. We did not consider studies containing colon or stomach slides, as they might already include some of the anomalies we aim to predict. We randomly sampled up to 50 slides per study and preprocessed the slides as described in the Methods section. The resulting dataset statistics are presented in Table \ref{tab:tcga-outlier-exposure-dataset}. We then trained models to discriminate between normal colon or stomach patches on one side and the TCGA OE patches on the other side and evaluated them on the held-out Charité cohort. In comparison to using natural images as OE, this improved the AD performance by a large margin (see Table \ref{tab:oe-dataset-ablation}).

\subsubsection*{OE using separately collected Charité data}
While TCGA data is closer to histopathology images than natural images, there are significant differences in how labs process slides. Also, TCGA data is limited in the diversity of patterns, as it mostly consists of cancerous tissue.
Therefore, we collected a diverse set of OE slides from Charité hospital and preprocessed them in the same manner as our frequent findings in the Charité cohort. This mitigates the model's ability to leverage certain low-level features, such as staining or resolution, in order to differentiate OE from normal data. The resulting \emph{Charité OE mixed} dataset consists of 2901 slides. We observed a significant performance improvement over the TCGA OE dataset (see Table \ref{tab:oe-dataset-ablation}), underscoring the significance of the OE data distribution being closely similar to that of the normal data.

\subsubsection*{Diversity of morphological patterns}
The observations in the previous sections showed that the composition of the OE data is crucial for strong generalization to true anomalies. We hypothesized that presenting the model with diverse morphological patterns helps to tighten the decision boundary around the normal data. Therefore, we evaluted different scenarios with varying diversity of OE data. The \emph{Kidney subset} and \emph{Breast subset} of \emph{Charité OE mixed} are limited in morphological diversity, as they contain only one tissue type. As expected, these limited subsets were outperformed by the more diverse complete \emph{Charité OE mixed} dataset. Interestingly, however, single tissue types that are morphologically very similar to the normal data (seperatly collected slides of small intestine biopsies, colon for stomach being normal, stomach for colon being normal) were competitive with the \emph{Charité OE mixed} performance. Combining all tissue types (\emph{Full OE set}) yielded the overall strongest performance.

\subsubsection*{Sampling}
The previous section has shown that both the diversity of OE Data and their similarity to the normal data are important for a well-generalizing AD model. Therefore, we sampled tissue that is similar and tissue that is more diverse equally in our approach. We defined tissue that is similar to the normal data as small intestine + colon or stomach and all other tissue types as diverse tissue types. From both sets, we sampled the OE data with equal probability.

\subsection{Data augmentation}  \label{app:augmentations}

The appearance of histopathology slides can vary considerably across stain manufacturers, scanners, and storage times \cite{macenko2009stainnorm, reinhard2001stainnorm, histo-xai-review}. To not overfit on the slide characteristics of the training set, we use stain normalization as well as training- and test-time augmentations. In this section, we will evaluate how much influence each of the different mechanisms has on the model performance.

\subsubsection*{Training augmentations}
For training purposes, we applied a set of data augmentations. We used random resized crops, which generate a crop in the range of 10\% to 100\% while maintaining $0.75$ of the image aspect ratio. Further, we used color jitter and transformed the image to grayscale with a probability of 20\%. The effect of different training augmentations is shown in Table \ref{tab:augmentation-ablation}. In all scenarios below, we use stain normalization as a preprocessing step.

\begin{table*}[h!]
    \caption{AD performance for OE with different forms of training augmentations for colon and stomach in the Charité and LMU cohorts.}
    \small
    \label{tab:augmentation-ablation}
    \centering
    \begin{tabular}{llcccc}
    \toprule
    & & \multicolumn{2}{|c|}{\textbf{Colon}} &  \multicolumn{2}{c}{\textbf{Stomach}} \\
    Cohort & Augmentation & \multicolumn{1}{|c}{patch-AUROC} & \multicolumn{1}{c|}{slide-AUROC} & patch-AUROC & slide-AUROC \\
    \midrule
    Charité &                         Crop Augmentations &           89.23 &           91.16 &             90.18 &             95.77 \\
        &                     Weaker Augmentations &           90.25 &           91.55 &             91.19 &             95.02 \\
        &                     Strong Augmentations &           90.38 &           91.49 &              90.90 &              94.40 \\
        \midrule
    LMU &                         Crop Augmentations &               - &           82.57 &                 - &             90.51 \\
        &                     Weaker Augmentations &               - &            85.60 &                 - &             91.42 \\
        &                     Strong Augmentations &               - &           84.58 &                 - &             93.07 \\
\bottomrule
    \end{tabular}
\end{table*}

We observed that data augmentations resulted in small performance improvements on the Charité cohort. However, on the LMU cohort, we saw larger improvements from color augmentations.
This indicates that the augmentations improve the models robustness to distribution shifts and also highlights the need to evaluate the models on a hold-out test cohort from a different hospital.

\subsubsection*{Test-time augmentation}
We used test-time augmentations to make our anomaly score estimation more robust. In detail, we augmented each image $k$ times, computed an anomaly score with our model, and then averaged the result. Table \ref{tab:test-time-aug} shows the effect of test-time augmentation on the Charité and the LMU cohort. 

\begin{table*}[h!]
    \caption{The effect of test-time augmentations on the performance of the Charité and LMU cohort.}
    \small
    \label{tab:test-time-aug}
    \centering
\begin{tabular}{llcccc}
\toprule
 & & \multicolumn{2}{|c|}{\textbf{Colon}} &  \multicolumn{2}{c}{\textbf{Stomach}} \\
    Cohort & Augmentation & \multicolumn{1}{|c}{patch-AUROC} & \multicolumn{1}{c|}{slide-AUROC} & patch-AUROC & slide-AUROC \\
\midrule
Charité & No Test-time augmentations &           89.72 &           91.09 &             90.74 &             94.66 \\
        &            5 Augmentations &           90.32 &           91.54 &             90.87 &             94.47 \\
        &           10 Augmentations &           90.38 &           91.49 &              90.90 &              94.40 \\
        \midrule
    LMU & No Test-time augmentations &               - &           82.24 &                 - &             87.59 \\
        &            5 Augmentations &               - &           84.77 &                 - &             92.49 \\
        &           10 Augmentations &               - &           84.58 &                 - &             93.07 \\
\bottomrule
\end{tabular}
\end{table*}

We observed that, on the Charité cohort, the performance with more test-time augmentations only marginally increased. However, for the LMU cohort, the performance notably increased from using no test-time augmentations to averaging the result of 5 augmentations. We used 10 augmentations for our results in the main paper.

\subsection{Model variations}  \label{app:model_variations}
In this section, we present an ablation on the model architecture and objective function of our AD model.

\subsection*{Network}
We compared fine-tuning a pretrained model (CTranspath \cite{ctranspath}) vs.\ training a Resnet-18 \cite{he2016deep} model from scratch on the OE task. Table \ref{tab:network-ablation} shows results on both cohorts in the 5-fold cross-validation setting from the main paper.

\begin{table*}[tbh]
    \caption{AD performance for OE with different network architectures for colon and stomach in the Charité and LMU cohorts. }
    \small
    \label{tab:network-ablation}
    \centering
    \begin{tabular}{llcccc}\toprule
    & & \multicolumn{2}{|c|}{\textbf{Colon}} &  \multicolumn{2}{c}{\textbf{Stomach}} \\
    Cohort & Architecture & \multicolumn{1}{|c}{slide-AUROC} & \multicolumn{1}{c|}{patch-AUROC} & slide-AUROC & patch-AUROC \\
    \midrule
Charité & ResNet-18 \cite{he2016deep} &                                       $90.57$ &         $90.3$ & $95.11$ &        $91.26$ \\
&  CTransPath \cite{ctranspath} &                                       $91.01$ &        $90.47$ & $95.04$ &        $91.37$ \\
\midrule
LMU & ResNet-18 \cite{he2016deep} &                                      $86.37$ &                  $-$ & $92.96$ &                  $-$        \\
& CTransPath \cite{ctranspath} &                                      $85.88$ &                  $-$ & $94.5$ &                  $-$         \\
\bottomrule
    \end{tabular}
\end{table*}

We observed that a pretrained model is not needed for a competitive anomaly detection model. The outlier task seems sufficient for the model to learn suitable representations that generalize from auxiliary anomalies to true anomalies.

\subsubsection*{Loss function}

Previous work has found that with OE, a simple binary cross-entropy loss outperforms specialized AD losses like DeepSAD \cite{ruff2020dsad} and HSC \cite{liznerski2022exposing}, which derive anomaly scores directly from the latent space. However, the authors attribute improved robustness to HSC, indicating superior performance in scenarios with  limited data availability or suboptimal OE samples. Hence, we assessed whether DeepSAD or HSC could improve the AD performance on our colon and stomach cohorts. Table \ref{tab:loss-function-ablation} shows the respective patch-AUROC and slide-AUROC scores on the Charité cohort. Neither method resulted in significant performance gains.

\begin{table*}[tbh]
    \caption{AD performance for OE with different loss functions in the colon and stomach Charité cohorts.}
    \label{tab:loss-function-ablation}
    \small
    \centering
    \begin{tabular}{lcccc}\toprule
    & \multicolumn{2}{c|}{\textbf{Colon}} &  \multicolumn{2}{c}{\textbf{Stomach}} \\
    Loss function & patch-AUROC & \multicolumn{1}{c|}{slide-AUROC} & patch-AUROC & slide-AUROC \\
    \midrule
         OE w/ BCE &           90.38 &           91.49 &              90.90 &              94.40 \\
             OE w/ HSC &           89.09 &           91.55 &             89.14 &              93.40 \\
         OE w/ DeepSAD &           89.11 &           91.66 &              89.30 &              94.60 \\
    \bottomrule
    \end{tabular}
\end{table*}

\noindent
In summary, our ablations show that a sufficiently high amount of training data as well as a fitting OE dataset are far more important for a successful AD model than the choice of network architecture or objective function.

\clearpage

\end{document}